\documentclass{article}

\PassOptionsToPackage{authoryear, compress}{natbib}

\usepackage[final]{neurips_2025}

\usepackage{graphicx}
\usepackage{makecell}
\usepackage{multirow}
\usepackage{amsmath}
\usepackage{amsthm}
\usepackage{soul}
\usepackage{fancybox}
\usepackage{tcolorbox}
\usepackage{subfigure}
\usepackage{algorithm}
\usepackage{algpseudocode}
\usepackage{bm}

\usepackage[utf8]{inputenc} 
\usepackage[T1]{fontenc}    
\usepackage{hyperref}       
\usepackage{url}            
\usepackage{booktabs}       
\usepackage{amsfonts}       
\usepackage{nicefrac}       
\usepackage{microtype}      
\usepackage{xcolor}         
\usepackage{wrapfig}

\title{Steering When Necessary: Flexible Steering Large Language Models with Backtracking}

\author{%
  Zifeng Cheng\footnotemark[1] , Jinwei Gan\footnotemark[1] , Zhiwei Jiang\footnotemark[2] , \\
  \textbf{Cong Wang}, \textbf{Yafeng Yin}, \textbf{Xiang Luo}, \textbf{Yuchen Fu}, \textbf{Qing Gu}\\
State Key Laboratory for Novel Software Technology, Nanjing University, China \\
  \texttt{chengzf@nju.edu.cn}, \texttt{ganjw@smail.nju.edu.cn}, \texttt{jzw@nju.edu.cn}, \\
  \texttt{cw@smail.nju.edu.cn}, \texttt{yafeng@nju.edu.cn}, \\
  \texttt{\{luoxiang,yuchenfu\}@smail.nju.edu.cn}, \texttt{guq@.nju.edu.cn}
}
\renewcommand{\thefootnote}{\fnsymbol{footnote}}

\begin{document}

\maketitle

\begin{abstract}
Large language models (LLMs) have achieved remarkable performance across many generation tasks.
Nevertheless, effectively aligning them with desired behaviors remains a significant challenge.
Activation steering is an effective and cost-efficient approach that directly modifies the activations of LLMs during the inference stage, aligning their responses with the desired behaviors and avoiding the high cost of fine-tuning.
Existing methods typically indiscriminately intervene to all generations or rely solely on the question to determine intervention, which limits the accurate assessment of the intervention strength.
To this end, we propose the \textbf{F}lexible \textbf{A}ctivation \textbf{S}teering with \textbf{B}acktracking (\textbf{FASB}) framework, which dynamically determines both the necessity and strength of intervention by tracking the internal states of the LLMs during generation, considering both the question and the generated content.
Since intervening after detecting a deviation from the desired behavior is often too late, we further propose the backtracking mechanism to correct the deviated tokens and steer the LLMs toward the desired behavior.
Extensive experiments on the TruthfulQA dataset and six multiple-choice datasets demonstrate that our method outperforms baselines.
Our code will be released at \url{https://github.com/gjw185/FASB}.
\end{abstract}

\footnotetext[1]{Equal Contribution.}
\footnotetext[2]{Corresponding Author.}
\renewcommand{\thefootnote}{\arabic{footnote}}

\section{Introduction}
\begin{wrapfigure}{r}{0.26\textwidth}
    \centering
    \includegraphics[width=0.26\textwidth]{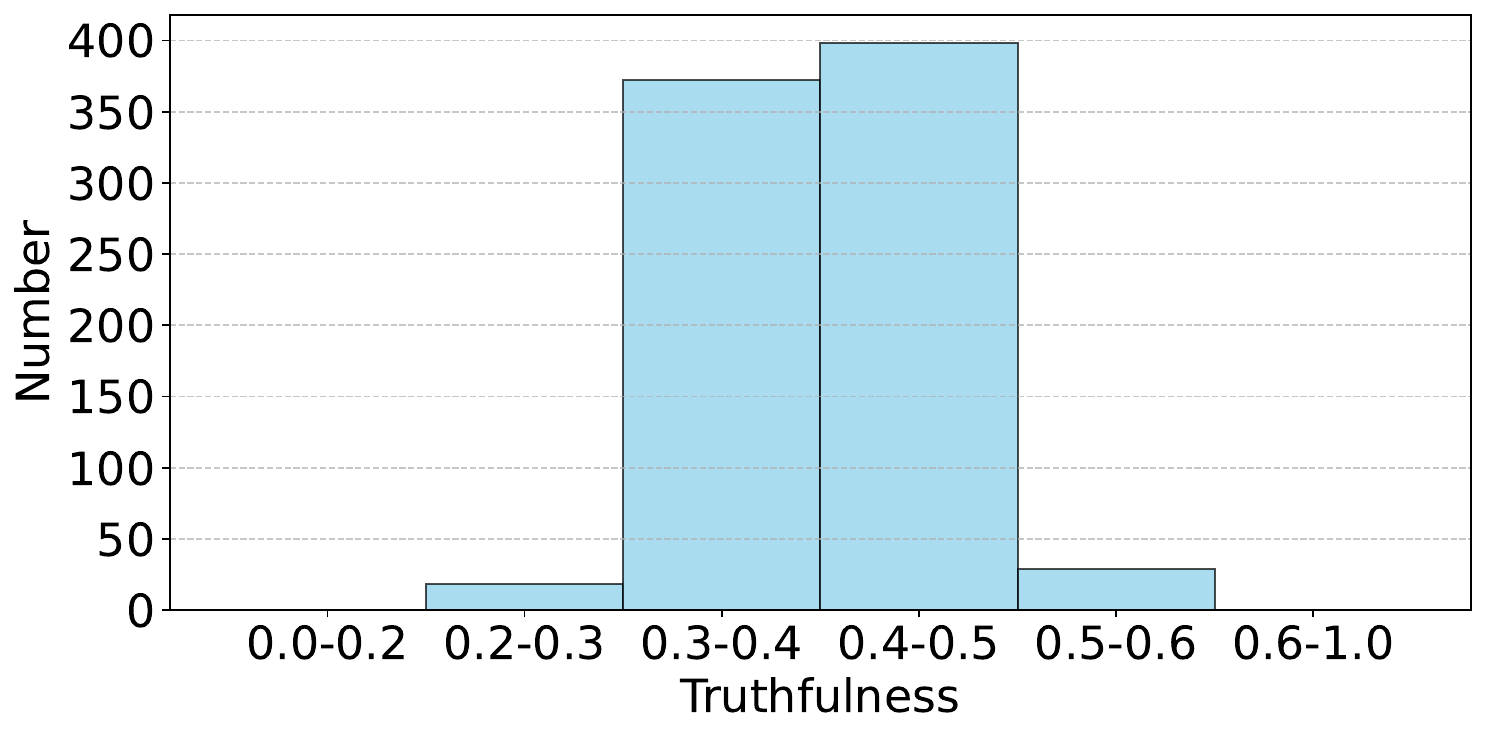}
    \caption{The truthfulness probability distribution of questions in the TruthfulQA dataset as detected by the classifier.}
    \label{fig:motivation}
\end{wrapfigure}
Large language models (LLMs)~\citep{LLAMA2,gpt3} have achieved great success in text generation.
However, the generated text still contains harmful information, hallucinations, and other misleading content.
Therefore, controlling LLMs to produce trustworthy, reliable, and other desired outputs remains a challenge.
Existing methods often use instruction tuning~\citep{DBLP:conf/iclr/WeiBZGYLDDL22}, Reinforcement Learning from Human Feedback (RLHF)~\citep{rlhf}, and prompt engineering~\citep{gpt3} to control LLMs.
Unfortunately, these methods often require large-scale datasets and expensive fine-tuning costs to achieve desired results.

Recently, activation steering or representation engineering~\citep{RE,DBLP:conf/acl/RimskyGSTHT24,DBLP:conf/nips/0002PVPW23,Turner,DBLP:conf/acl/ChengWFJ00025} has been proposed to control the outputs of LLMs by directly modifying their internal activations during inference, thereby avoiding the high cost associated with large-scale data collection and fine-tuning.
This technique constructs steering vectors from positive and negative samples to make targeted modifications to the activations of the LLM, enabling more precise control over its output.

\begin{figure*}[t]
  \centering
  \includegraphics[width=\textwidth]{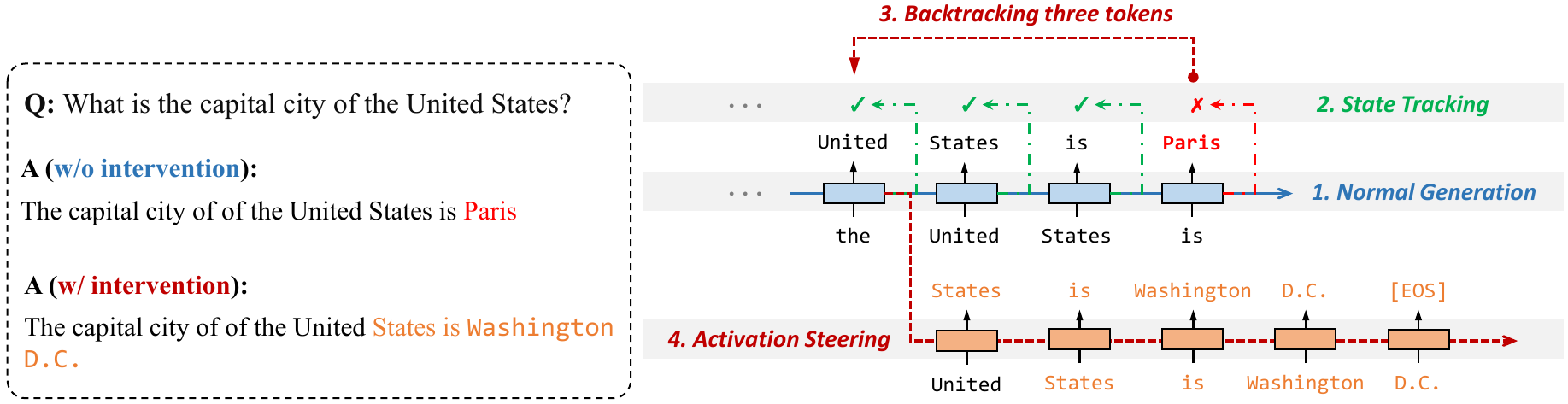}
  \caption{
  The overview of flexible activation steering with backtracking framework.}
  \label{fig:overview}
\end{figure*}

Existing methods often apply interventions indiscriminately to all generations~\citep{DBLP:conf/acl/RimskyGSTHT24,sadi,DBLP:conf/nips/0002PVPW23} or determine whether and how strongly to intervene based solely on the question~\citep{DBLP:conf/iclr/LeePRMDND25}, which limits the accurate assessment of the intervention strength.
First, different generations should receive different levels of intervention. If a generation does not deviate, no intervention is necessary. If it deviates significantly, a stronger intervention should be applied.
For example, when LLM is asked ``Where is the capital of the United States?''.
If the response is ``The capital of the United States is Washington, D.C.'', no intervention is necessary, as the answer is correct.
If the response is ``Paris'', it often requires a stronger intervention compared to ``New York''.
Second, as each LLM generates different responses to the same question, question-only probing is challenging\footnote{Questions in some domains may be relatively simple and general, such as those related to safety.} and requires data collection for each LLM, resulting in high overhead.
On the other hand, training a classifier on concatenated questions and answers often struggles to directly probe the question to determine whether and how strongly to intervene.
As shown in Figure \ref{fig:motivation}, the predicted truthfulness probabilities for questions in the TruthfulQA dataset are concentrated between 0.3 and 0.5, making fine-grained decisions difficult.

In this paper, we propose a \textbf{F}lexible \textbf{A}ctivation \textbf{S}teering with \textbf{B}acktracking (\textbf{FASB}) method, as shown in Figure \ref{fig:overview}.
Unlike existing methods, \textbf{FASB} tracks the internal states of the LLM after each normal generation step, taking both the question and the generated content into account.
In this way, \textbf{FASB} can dynamically determine whether intervention is necessary and the intervention strength based on the degree of deviation in the generation.
Specifically, FASB employs two methods to identify internal states that are consistent with the desired behavior and derive the steering vector and classifier for state tracking.
Considering that intervening after detecting a deviation from the desired behavior is often too late, we further propose the backtracking mechanism.
The backtracking mechanism steps back a few tokens and performs activation steering to regenerate them, steering the generation toward the desired direction, with the intervention strength determined by the classifier.


Our main contributions are as follows:
\begin{itemize}
\item We propose a flexible activation steering with backtracking that dynamically determines both the necessity and strength of intervention by tracking the internal states of the LLMs.
\item We propose a backtracking mechanism to step back a few tokens in order to apply intervention and regenerate them.
\item We conduct extensive experiments on the TruthfulQA dataset and six multiple-choice datasets to demonstrate the effectiveness of our method.
\end{itemize}

\section{Related Work}
Activation steering or representation engineering~\citep{RE,DBLP:conf/acl/RimskyGSTHT24,DBLP:conf/nips/0002PVPW23,Turner,leong2023self,wang2025advanced,wangensembling} uses steering vectors to directly modify the activations of LLMs during inference to control their outputs in a desired direction.
Activation steering preserves the general capabilities of LLMs while avoiding the expensive costs of high-quality data collection and fine-tuning~\citep{DBLP:conf/acl/ZhangZYZGCW25,DBLP:conf/acl/ZhongZZ25,shen2025imagdressing}.

The pioneering work~\citep{RE,Turner,wang2024v} creates and compares two prompts to obtain the steering vector.
ITI~\citep{DBLP:conf/nips/0002PVPW23} uses the probe tool on a contrastive question-answering dataset to identify a set of attention heads associated with truthfulness.
During inference, ITI modifies the activations of all subsequent generations in directions associated with truthfulness.
Truth Forest~\citep{TruthForest} further employs multiple orthogonal probes to extract several truthful steering vectors, thereby improving performance.
ACT~\citep{act} uses clustering to construct multiple steering vectors and proposes an adaptive steering strength. 
CAA~\citep{DBLP:conf/acl/RimskyGSTHT24} computes steering vectors by averaging the difference between pairs of positive and negative examples. During inference, these steering vectors are added into the residual stream with a chosen steering strength at all token positions after the prompt to control the direction.
ORTHO~\citep{refusal} also averages the difference between pairs of positive and negative to compute the steering vector and performs directional ablation using the opposite direction of the steering vector to guide the model toward the desired behavior.
SADI~\citep{sadi} utilizes activation differences in contrastive pairs to precisely identify intervention position and dynamically steers model behavior by scaling element-wise activations.
In addition, some works have explored the use of activation steering in instruction-following~\citep{Instruction-Following}, in-context learning~\citep{DBLP:conf/icml/LiuY0Z24}, and differentially private~\citep{private}.

\begin{figure*}[t]
  \centering
  \includegraphics[width=\linewidth,trim=3.3cm 1.1cm 5.1cm 2.2cm,clip]{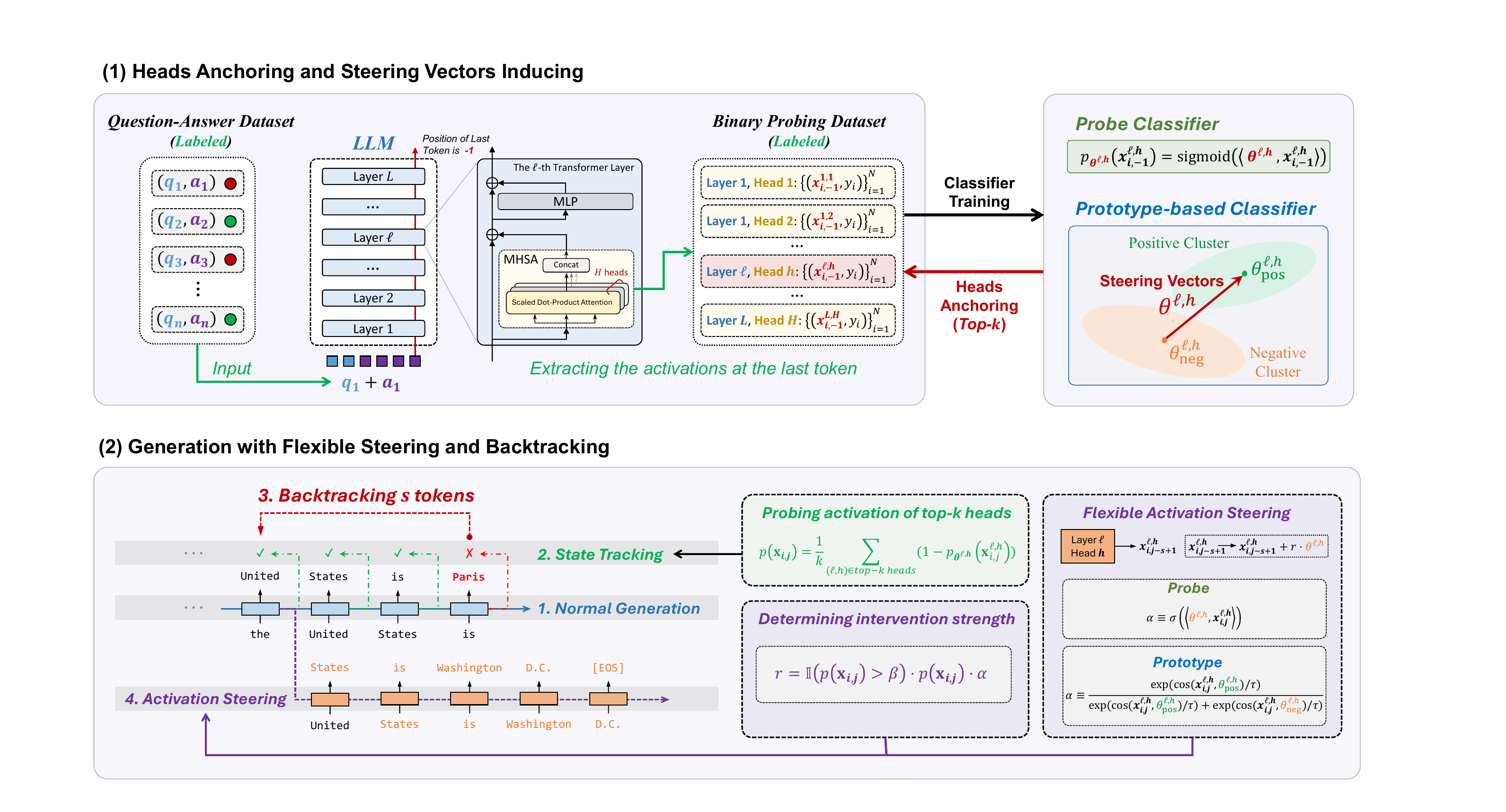}
  \caption{
  The framework of flexible activation steering with backtracking.}
  \label{fig:framework}
\end{figure*}

\section{Method}
Our method consists of two steps to flexibly steer model behavior, as illustrated in Figure~\ref{fig:framework}.
In the first step, we employ two alternative methods to identify the attention heads for intervention and to derive both the steering vector and the classifier.
In the second step, the classifier is used for state tracking to determine whether intervention is necessary and to adaptively set intervention strength.
We further propose a backtracking mechanism that allows the LLMs to regenerate tokens that deviate from the desired behavior.
The full procedure can be found in Algorithm \ref{algorithm} in the Appendix.

\subsection{Heads Anchoring and Steering Vectors Inducing}
The first step is to use the classifier to identify the attention heads related to the desired behavior and to construct the steering vector.
We use the \textbf{Probe} method to identify attention heads and induce the classifiers and steering vectors. 
In the Appendix \ref{app:A}, we present an alternative \textbf{Prototype} method.
Since the \textbf{Probe} method uses a trainable classifier and typically achieves higher accuracy, we focus only on the \textbf{Probe} method in the section.


\paragraph{Probe} Probe method employs standard probing tools to identify attention heads.
The idea behind probing tools is to train a lightweight classifier (probe) on the activations of attention heads to identify the relevant heads.
Subsequently, we select the top-$k$ heads with the highest accuracy on the validation set for intervention, as they can effectively separate the samples and are more aligned with the desired behavior.

Specifically, we concatenate the question and answer from the labeled dataset and extract the activations at the last token to collect a binary probing dataset $\{ (\textbf{x}_{i,\text{-}1}^{\ell,h}, y_i) \}_{i=1}^N$ for each head in each layer, where $\textbf{x}_{i,\text{-}1}^{\ell,h}$ denotes the activation of $h$-th attention head in the $\ell$-layer at the last token, and $y_i \in \{0, 1\}$ denotes the label.
The probe takes the following form:
\begin{equation}
    p_{\theta^{\ell,h}}(\textbf{x}_{i,\text{-}1}^{\ell,h}) = \text{sigmoid}(\langle \theta^{\ell,h}, \textbf{x}_{i,\text{-}1}^{\ell,h}\rangle) \label{probe:classifier}
\end{equation}
where $\langle,\rangle$ denotes dot product, and $\theta^{\ell,h}$ denotes the probe parameter of $h$-th attention head in the $\ell$-layer.
Since the probes can effectively separate positive and negative examples, we use the parameters of probe $\theta^{\ell,h}$ as the steering vector.

\begin{figure*}
  \centering
  \subfigure[Accuracy of Probe]{
        \includegraphics [width=0.47\textwidth,trim=0.2cm 0.25cm 0.2cm 0.2cm,clip]{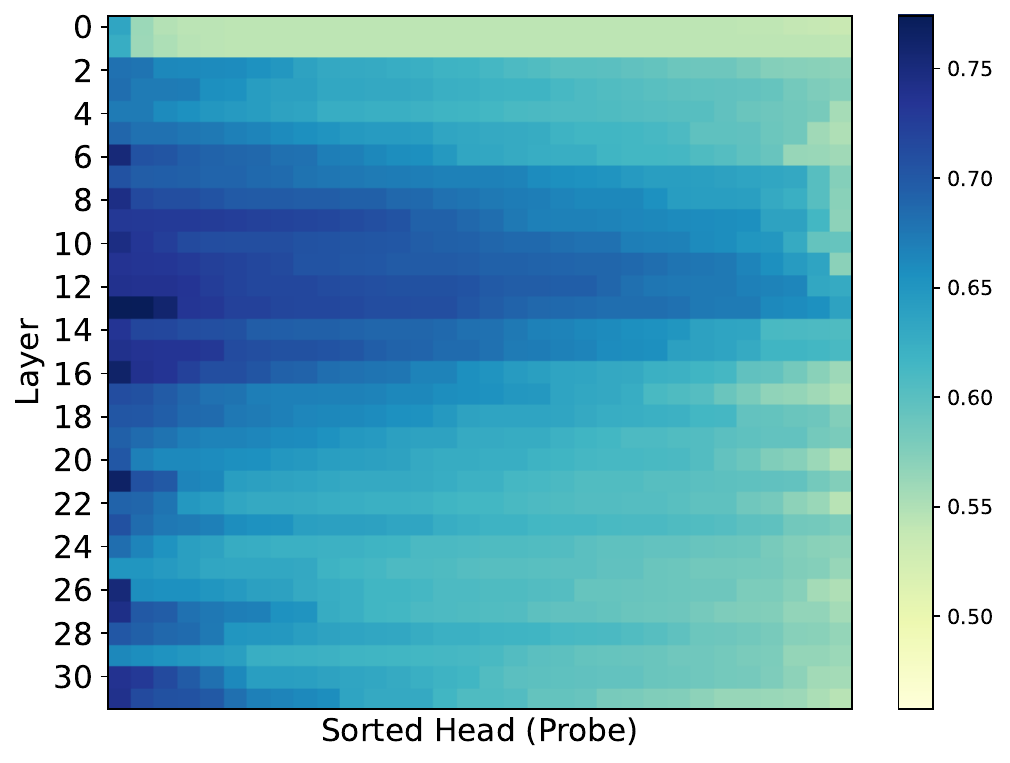}
        }
        \subfigure[Accuracy of Prototype]{
 \includegraphics [width=0.47 \textwidth,trim=0.2cm 0.25cm 0.2cm 0.2cm,clip]{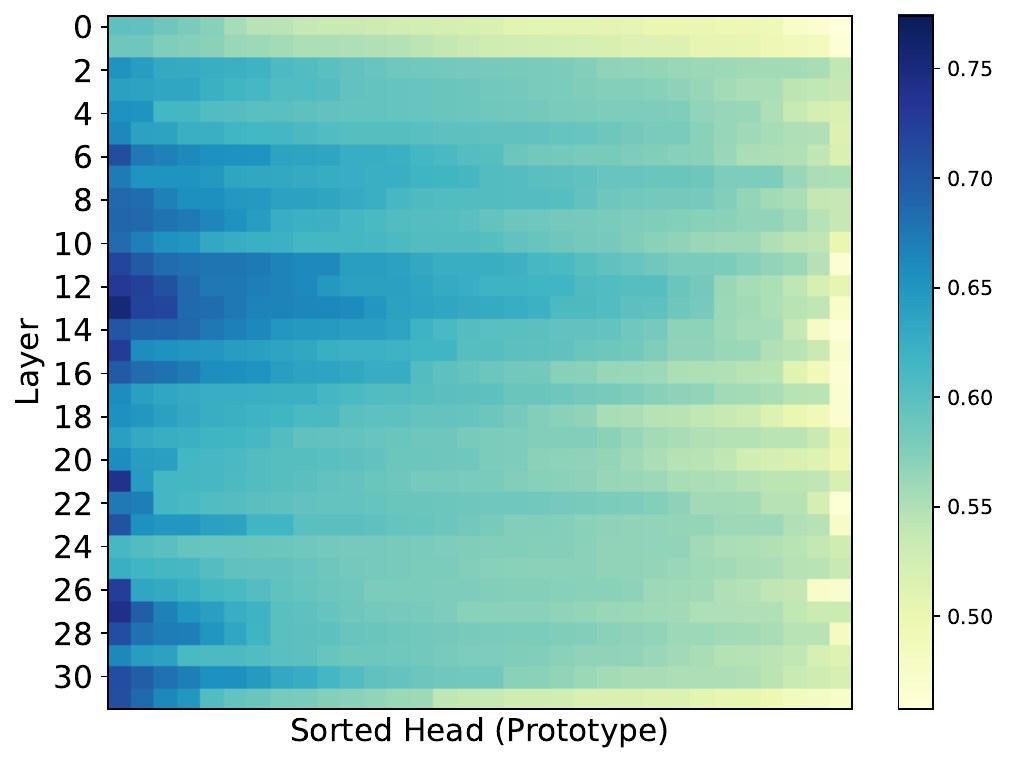}
  }
  \caption{Accuracies on the validation set of TruthfulQA dataset for all heads in all layers in LLaMA2-7B-CHAT, sorted row-wise by accuracy. Darker blue represents higher accuracy.} \label{acc}
\end{figure*}

As shown in Figure~\ref{acc}, a subset of heads is strongly related to the truthfulness, and the optimal heads are relatively evenly distributed across all layers.
Therefore, performing fine-grained intervention only on these heads is more targeted and helps minimize disruption to the model's overall behavior.
It is worth noting that our method can also intervene in the outputs of the MLP module or the outputs of layers.

\subsection{Generation with Flexible Steering and Backtracking}
In the second step, after generating each token, the classifier is used for state tracking to determine whether the generation deviates from the desired behavior, allowing for flexible intervention.
If a deviation is detected during the generation process, it often indicates that the previously generated tokens have deviated from the desired behavior.
We backtrack a few tokens to regenerate them and apply steering vectors to adaptively intervene in all subsequent tokens.
The backtracking mechanism enables the correction of deviated tokens and helps steer the model toward the desired behavior.
If no deviation is detected, the model continues generating normally.

\paragraph{State Tracking}
After generating the $j$-th token, the classifier probes the $j$-th token's activation of the LLM to assess whether the current generation deviates from the desired behavior and determines the intervention strength.
Notably, state tracking and probe share the same goal: determining whether the current response deviates from the desired behavior.
Specifically, we average the prediction probabilities from the top-$k$ selected heads to assess the deviation:
\begin{equation}
    p(\textbf{x}_{i,j}) = \frac{1}{k} \sum_{(\ell,h) \in \text{top-$k$ heads}} (1 - p_{\theta^{\ell,h}}(\textbf{x}_{i,j}^{\ell,h}))  \label{p}
\end{equation}
where $p(\textbf{x}_{i,j})$ denotes the deviation probability after generating $j$-th token for $\textbf{x}_i$, and $\textbf{x}_{i,j}^{\ell,h}$ represents the activation of $j$-th generated token for $\textbf{x}_i$ at the $\ell$-th layer and the $h$-th head.
The process is highly lightweight, as the hidden states are already generated during the generation process.

\paragraph{Backtracking}
When the deviation probability exceeds the threshold, we consider the generation to deviate from the desired behavior, and intervention should be applied.
An important issue is that intervening only after a deviation is detected is often too late, as it can only affect the content generated subsequently.
To address this issue, we propose a backtracking mechanism to regenerate the previous deviated tokens.
If a deviation is detected after generating the $j$-th token, the backtracking mechanism retains only the first $j$-$s$ tokens and regenerates the subsequent tokens, where $s$ is a hyperparameter that controls the number of tokens to backtrack.
Compared to generating from scratch, the backtracking mechanism only requires generating an additional $s$ tokens, making the overhead lightweight.
Subsequently, we apply intervention and regenerate the content after the ($j$-$s$)-th token to prevent further deviation.

\paragraph{Activation Steering}
We further introduce an adaptive intervention strength for activation steering, where the strength is determined by the degree of deviation calculated by the classifier and is proportional to it.
The intervention strength is defined as follows:
\begin{equation}
    r = \mathbb{I}(p(\textbf{x}_{i,j}) > \beta) \cdot p(\textbf{x}_{i,j}) \cdot \alpha
\label{strength}
\end{equation}
where $\alpha$ and $\beta$ are hyperparameters that control the intervention strength and the threshold for deviation probability, respectively.
An appropriately sized $\alpha$ can steer the LLM toward the desired behavior without compromising generation quality.
When $\alpha$ is too large, it degrades generation quality, whereas when it is too small, it fails to provide effective steering.
An appropriately sized $\beta$ allows intervention when deviations occur.
An excessively large $\beta$ may miss necessary interventions, while an excessively small one may cause over-intervention.

Since we intervene in the output of selected heads, the MHSA with intervention can be formulated as follows:
\begin{align}
    \textbf{h}^{\ell}_{i,j\text{-}s\text{+}1} =  \text{concat}(\textbf{x}^{\ell,1}_{i,j\text{-}s\text{+}1} + r \theta^{\ell,1} c^{\ell,1}, \cdots, \textbf{x}^{\ell,H}_{i,j\text{-}s\text{+}1} + r \theta^{\ell,H} c^{\ell,H}) \textbf{W}^{\ell,O}  \label{steering}
\end{align}
where $\textbf{h}^{\ell}_{i,j\text{-}s\text{+}1}$ represents the output of $\ell$-th MHSA at the ($j\text{-}s\text{+}1$)-th token, $\textbf{x}^{\ell,H}_{i,j\text{-}s\text{+}1}$ represents the output of self-attention for $H$-th head in $\ell$-th layer, $\textbf{W}^{\ell,O}$ is output projection matrix in $\ell$-th layer.
$c^{\ell,H}$ is a binary scalar that equals 1 for the selected top-$k$ heads and 0 for the unselected ones.

\section{Experiments}
\subsection{Experimental Settings}
\textbf{Datasets and Evaluation Metrics}
We conduct experiments on open-ended generation tasks and multiple-choice tasks.
TruthfulQA~\citep{truthfulqa} dataset includes open-ended generation task and multiple-choice task.
For the open-ended generation, we employ two fine-tuned LLMs to judge whether the answer is truthful\footnote{https://huggingface.co/allenai/truthfulqa-truth-judge-llama2-7B} and informative\footnote{https://huggingface.co/allenai/truthfulqa-info-judge-llama2-7B}, denoted as True (\%) and Info (\%) respectively, while the product True*Info (\%) serves as the primary metric.
For the multiple-choice tasks, we use datasets: COPA~\citep{copa}, StoryCloze~\citep{story}, NLI~\citep{nli}, MMLU~\citep{mmlu}, SST2~\citep{sst2}, and Winogrande~\citep{WinoGrande}, with response formats ranging from 2-way to 4-way choices.
We use multiple-choice accuracy (MC) to evaluate.

\textbf{Implementation Details}
In the Probe method, for the TruthfulQA dataset, we intervene using the top-24 heads, set the threshold range to [0.4, 0.5], the number of backtracking steps to 10, and search for the intervention strength in the range of [40, 80] with a step size of 10.
For the six multiple-choice tasks, our threshold search range is [0.3, 0.4, 0.5, 0.6], the intervention strength search range is [0, 250] with a step size of 10, and the number of backtracking steps is 10.

\begin{table}[t]
\centering
\small
\caption{Results on TruthfulQA open-ended generation (True*Info \%) and multiple-choice tasks (MC \%).} \label{tab:main}
\begin{tabular}{lcccccc}
\toprule
\multirow{2}{*}{\textbf{Methods}} & \multicolumn{3}{c}{\textbf{Open-ended Generation}} & \multicolumn{3}{c}{\textbf{Multiple-Choice}} \\ \cmidrule(lr){2-4} \cmidrule(lr){5-7}
& True (\%) & Info (\%) & True*Info (\%)   & MC1 (\%)   & MC2 (\%)  & MC3 (\%)      \\ 
\midrule
\textbf{Baseline}                 & 66.83       & 99.51       & 66.50            & 33.41         & 51.07         & 24.76        \\
\textbf{ITI}       & 94.49     & 80.55   & 76.11 & 38.31      & 57.15      & 30.53    \\
\textbf{CAA}   & 71.60 & 83.84 & 60.03 & 34.03 & 52.76 & 25.62 \\
\textbf{ORTHO}       & 67.94     & 90.09   & 61.21 & 36.23      & 52.88      & 26.12   \\
\textbf{CAST}       & 67.69     & 86.17   & 58.33 & 33.90      & 51.17      & 25.01   \\
\textbf{ACT}       &-      &-    & - & 28.80      & 45.20      & -   \\
\textbf{SADI-HEAD} & 77.72 & 98.53 & 76.58 & 35.90 & 54.65 & 26.99\\
\midrule
\textbf{Probe (Ours)}   & 93.88 & 85.81 & \textbf{80.56} & \textbf{48.71} & \textbf{66.58}   & \textbf{41.95} \\
\bottomrule
\end{tabular}
\end{table}

\begin{table}[t] \small
\centering
\caption{Results (MC \%) on six multiple-choice tasks.}
\label{overall-results-choice}
\begin{tabular}{lcccccccc}
\toprule
\textbf{Methods} & \textbf{COPA}  & \textbf{StoryCloze} & \textbf{NLI}    & \textbf{MMLU}   & \textbf{SST2}   & \textbf{Winogrande} & \textbf{\texttt{AVG}} \\ 
\midrule
\textbf{Baseline} & 64.4 & 60.2 & 63.5  & 60.2  & 92.2 & 50.2   & 65.1 \\ 
\textbf{ITI} & 66.6 & 59.7 & 64.3 & 60.2 & 92.3 & 51.5 & 65.8 \\
\textbf{CAA} &  66.6 & 63.5 & 64.9 & 62.6 & 92.2 & 50.9 & 66.8 \\
\textbf{ORTHO} & 65.2 & 60.2 & 63.1 & \textbf{63.8} & 92.4 & 50.4 & 65.8 \\
\textbf{SADI}   & 65.4   & 60.5  & 65.1 & 61.8 & 92.3 & 51.8 & 66.1 \\
\midrule
\textbf{Probe (Ours)} & \textbf{90.0} & \textbf{93.5} & \textbf{80.0} & 62.4 & \textbf{92.8} & \textbf{54.1} & \textbf{78.8} \\
\bottomrule
\end{tabular}
\end{table}

\subsection{Baselines}
We compare our model with the following baselines to show its effectiveness.
\textbf{Baseline} directly uses the original LLaMA2-7B-CHAT model to generate text.
\textbf{ITI}~\citep{DBLP:conf/nips/0002PVPW23} identifies a set of attention heads with high linear probing accuracy and shifts activations of all subsequent generations following the user's prompt along these probe-correlated directions.
\textbf{CAA}~\citep{DBLP:conf/acl/RimskyGSTHT24} computes the steering vector by averaging the difference between pairs of positive and negative examples. During inference, these steering vectors are added at all generations with a coefficient.
\textbf{ORTHO}~\citep{refusal} uses the same reversed steering vector for directional ablation, steering the generation toward the desired direction.
\textbf{CAST}~\citep{lee2024programming} computes the condition vector and behavior vectors using PCA.
During the inference phase, it uses the condition vector to make judgments, enabling dynamic intervention in generations.
\textbf{ACT}~\citep{act} constructs multiple classifiers through clustering and uses these classifiers to dynamically intervene in responses from different directions.
\textbf{SADI}~\citep{sadi} utilizes activation differences in contrastive pairs to precisely identify intervention position and dynamically steer model behavior by scaling element-wise activations.

\subsection{Results}

\textbf{Results on TruthfulQA} 
In Table~\ref{tab:main}, the Probe method demonstrates superior performance compared to the baselines on the TruthfulQA dataset.
Compared to the ITI method, our Probe method does not require additional training on top of ITI and has achieved performance far exceeding that of the ITI method.
This is because ITI applies the same intervention strength indiscriminately to all generations.
This makes it impossible to assign adaptive intervention strength and may cause originally correct answers to be incorrectly altered due to the intervention.
In contrast, our method adaptively determines the intervention strength and does not intervene in correct generations.
Compared with other dynamic intervention methods such as ORTHO, CAST, ACT, and SADI, it can be noted that SADI also achieves good performance, which demonstrates that the dynamic steering vector is effective.
ORTHO and CAST are mainly designed for safety-related scenarios, where it is relatively easy to determine whether a query is harmful based solely on the query information.
However, in domains such as truthfulness and faithfulness, it is difficult to anticipate whether the generated content will deviate based solely on the query, resulting in poor performance.
Compared with these dynamic intervention methods, the superiority of our method can be further demonstrated.


\textbf{Results on Multiple-choice} In Table~\ref{overall-results-choice}, our method achieves consistently promising results across all datasets, and Probe method achieves the best performance.
This demonstrates that adaptive intervention and using probes to select heads remain effective on multiple-choice datasets.
Notably, CAA achieves better results than ITI, suggesting that layer-wise intervention can sometimes lead to better performance.
SADI also performs well in multiple-choice tasks, which demonstrates the generalizability of the dynamic steering vector.


\begin{table}[t]  \setlength{\tabcolsep}{10pt}
\centering
\small
\caption{Ablation study on TruthfulQA dataset.}
\label{tab:ablation}
\begin{tabular}{lccc}
\toprule
\textbf{Methods} & \textbf{True*Info}   & \textbf{MC1}   & \textbf{MC2}  \\
\midrule
\textbf{Probe}   & \ 80.56 & \textbf{48.71} & \textbf{66.58}  \\
\midrule
\multicolumn{4}{l}{\it{Intervention Strength:}}\\
\textbf{All fixed strength} & 76.11 & \ 38.31 & \ 57.15 \\
\textbf{w/o Adaptive} & \textbf{82.08} & 42.96 & 62.06\\
\midrule
\multicolumn{4}{l}{\it{Intervention Position:}}\\
\textbf{w/o Backtracking} &  62.11 & \ 35.01 & \ 53.55 \\
\textbf{After Question} & 72.55 & \ 41.86 & \ 59.88\\
\bottomrule
\end{tabular}
\end{table}

\subsection{Ablation Study}
We conduct an ablation study to show the effectiveness of the proposed components in Table~\ref{tab:ablation}.

We first explore the ablation of intervention strength.
``All fixed strength'' refers to the results of applying the same intervention strength to all samples. 
``w/o Adaptive'' refers to the results of applying the same intervention strength to samples that meet the intervention criteria.
By comparing the results of ``w/o Adaptive'' with ``All fixed strength'', it is demonstrated that intervening only on samples that deviate from the desired behavior can improve performance.
Comparing the results of the Probe with ``w/o Adaptive'' demonstrates the effectiveness of dynamic intervention strength. 

We then explore the ablation of the intervention position.
``w/o Backtracking'' means no backtracking is performed.
``After Question'' indicates that the representation of the question was used to decide whether to intervene and intervention strength.
Comparing the results of the Probe with ``w/o Backtracking'' shows the necessity of the backtracking operation. 
The True*Info metric of ``w/o Backtracking'' is even lower than the baseline.
This indicates that intervening after detecting the deviation from the desired behavior is too late.
Comparing the results of the Probe with ``After Question'' indicates that relying solely on the hidden states of the question part to judge whether the subsequent response deviates from the desired behavior is insufficient.
It is necessary to make judgments after generating part of the response.

\subsection{Generalizability across More Truthful Benchmarks}
We further investigate whether our method can generalize beyond the TruthfulQA benchmark.
Specifically, we directly evaluate the classifier and steering vectors obtained from TruthfulQA on two datasets related to real-world truth, including Natural Questions~\citep{NQ} and TriviaQA~\citep{TriviaQA}.
The Natural Questions dataset consists of 3,610 real queries issued to the Google search engine, annotated with answers and supporting Wikipedia pages.
TriviaQA includes 95K question-answer pairs annotated by trivia enthusiasts.
Following~\citet{DBLP:conf/nips/0002PVPW23}, all benchmarks are presented in a multiple-choice format.

The results indicate that the Probe method outperforms the baseline and the ITI on both benchmarks, as shown in Table~\ref{tab:Generalizability}.
This suggests that employing the Probe method does not undermine the model's performance in out-of-distribution truthful domains; Instead, it enhances the model's performance, particularly in domains closely related to the real-world truth.
This indicates that the classifier and steering vector learned on the TruthfulQA dataset are not domain-specific, but rather general real-world truth features.

\begin{table}[t]
\centering
\small
\caption{MC2 on the Natural Questions and TriviaQA multiple-choice datasets using LLaMA2-7B-CHAT as the baseline.}
\label{tab:Generalizability}
\begin{tabular}{lcc}
\toprule
\textbf{Methods} & \textbf{Natural
Questions} & \textbf{TriviaQA}  \\
\midrule
\textbf{Baseline} & 49.54 & 61.22 \\
\textbf{ITI} & 56.50 & 66.49 \\
\textbf{Probe} & \textbf{59.25} & \textbf{67.55}  \\
\bottomrule
\end{tabular}
\end{table}

\begin{figure}[t]
  \centering
  \includegraphics[width=1\linewidth,trim=0.2cm 0.25cm 0.1cm 0.15cm,clip]{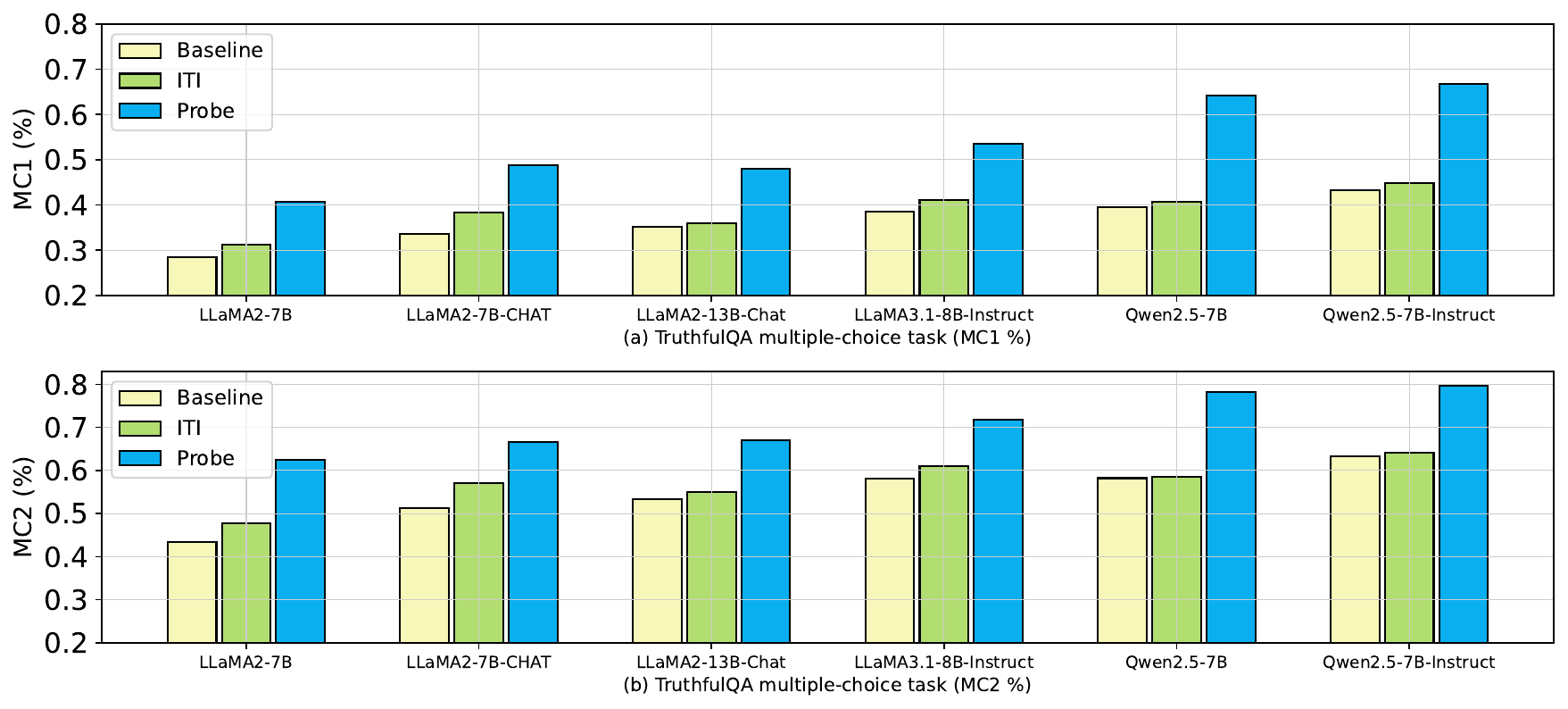}
  \caption{The performance of various LLMs on the TruthfulQA benchmark.}
  \label{fig:various_LLMs}
\end{figure}

\begin{figure*}[t]  \setlength{\tabcolsep}{8pt}
  \centering
  \includegraphics[width=1\linewidth,trim=0.2cm 0.2cm 0.15cm 0.2cm,clip]{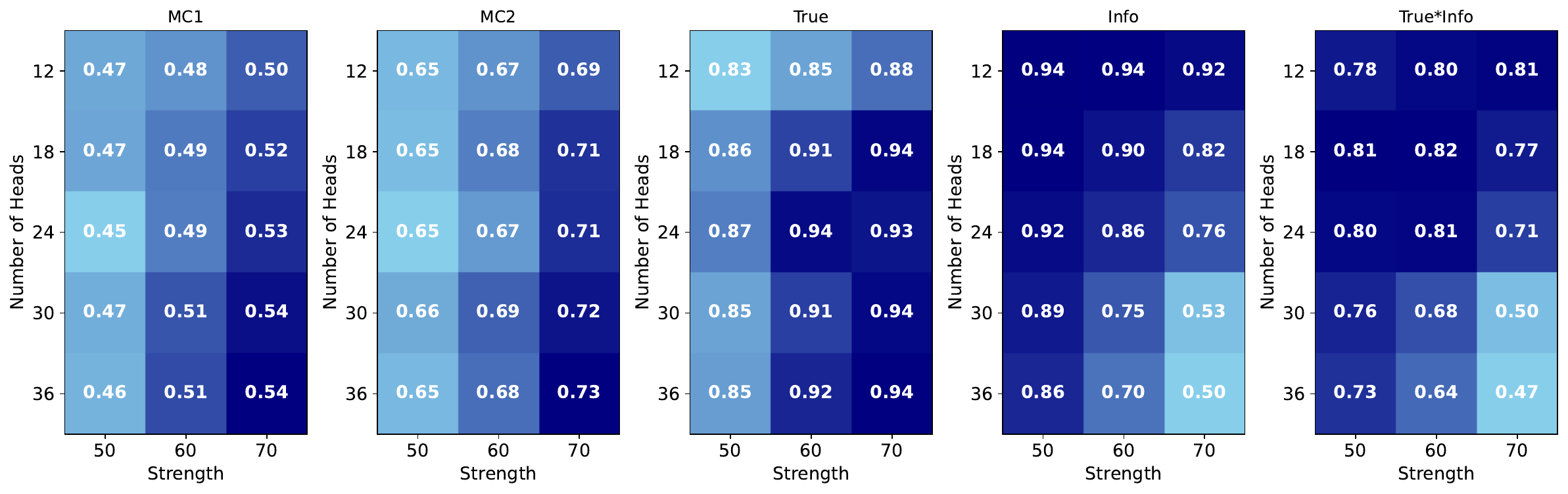}
  \caption{Results with varying intervention strength and numbers of attention heads on the TruthfulQA dataset with LLaMA2-7B-CHAT.}
  \label{fig:head_strength}
\end{figure*}

\subsection{Generality across More LLMs}
To evaluate the generality across various LLMs, we apply the Probe method to 6 LLMs, including LLaMA2~\cite{LLAMA2}, LLaMA3.1~\citep{llama3}, and Qwen2.5~\citep{qwen2.5}.
The performance of baseline, ITI, and Probe on the TruthfulQA benchmark in Figure~\ref{fig:various_LLMs}.

The probing method effectively improves performance across 6 LLMs, demonstrating that our approach can generalize to models of different sizes, architectures, and whether or not they have been instruction-tuned.
Notably, our method yields significant improvements of 24.61 and 20.03 in MC1 and MC2, respectively, on Qwen2.5-7B.
Compared with the ITI method, our method achieves better performance on all 6 LLMs.
On Qwen2.5-7B, there is an improvement of 23.50 on the MC1 and an improvement of 19.81 on the MC2.
It is demonstrated that our method outperforms the ITI method in terms of performance across different LLMs.
Scaling up the model size often results in better performance, indicating that it has acquired more knowledge related to truthfulness.
Instruction-tuned models achieve better performance than non-instruction-tuned ones, indicating that instruction tuning helps enhance the truthfulness of the model.

\subsection{Model Analysis}
\paragraph{Effects of intervention strength and head number}
In Figure~\ref{fig:head_strength}, we present the results of our method on the TruthfulQA dataset. 
We sweep two hyperparameters to control the intervention: the number of identified attention heads and the intervention strength. 

The truthful metrics MC1, MC2, and True show a trend where the results continuously improve as both the strength and the number of attention heads increase.
This is because, with the continuous increase in the number of heads and the strength of intervention, the internal truthfulness of the model is constantly increasing, making the model more inclined to choose and respond with more truthful answers.
It is worth noting that our method can be applied with greater strength, as it does not intervene with all samples, and the intervention strength is adaptive.

The Info metric decreases as the number of heads increases and the intervention strength increases.
This is because, with the continuous increase in intervention strength, the model internally encodes more truthfulness, leading the model to be more inclined to respond with answers that are truthful but lack informativeness.
Due to the mutual restraint between the True metric and the Info metric, the True*Info metric achieves highest values when the True metric and the Info metric are relatively balanced, i.e., the number of heads is 18 and the strength is 60.
Therefore, the optimal hyperparameters for MC2 and True*Info are not consistent.

\begin{figure}
  \centering
\subfigure[Effects of threshold]{
     \includegraphics 
     [width=0.48 \textwidth,trim=0.2cm 0.3cm 0.15cm 0.2cm,clip]{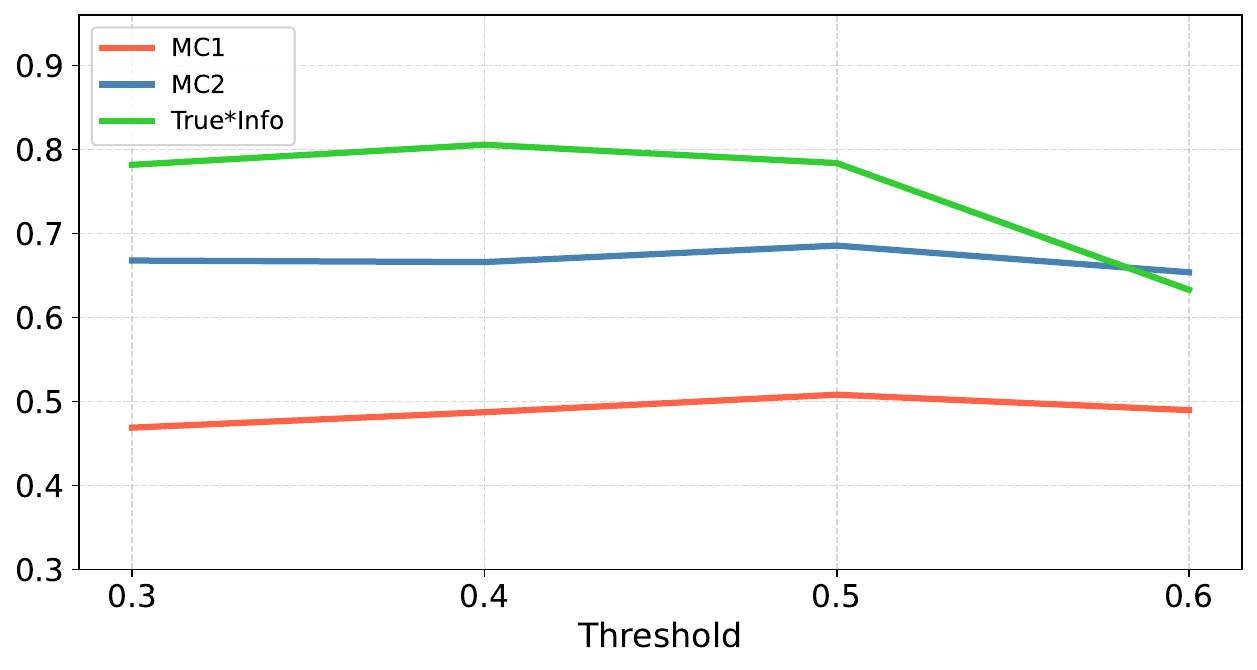}
      }
\subfigure[Effects of size of the training set]{
     \includegraphics 
     [width=0.48 \textwidth,trim=0.25cm 0.2cm 0.2cm 0.2cm,clip]{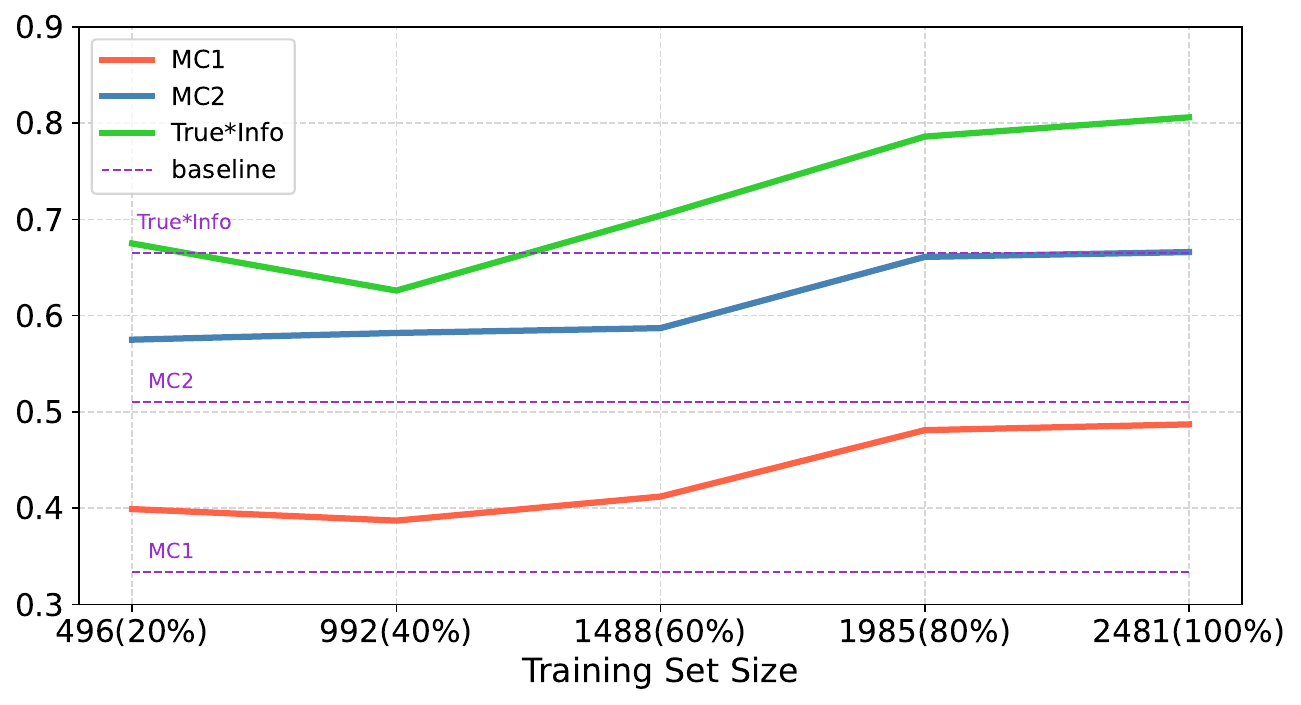}
      }
      \caption{Effects of the threshold and the size of the training set on TruthfulQA dataset.} \label{step_and_size}
\end{figure}

\paragraph{Effects of threshold}
To investigate the impact of different thresholds $\beta$, we conduct experiments with thresholds set at 0.3, 0.4, 0.5, and 0.6.
When $\beta$ is 0.3, 0.4, or 0.5, the overall performance does not vary significantly. Among them, True*Info achieves the best performance at 0.4, while MC1 and MC2 perform best at 0.5.
When the threshold is set to 0.6, True*Info drops sharply, which may be due to excessive intervention.

\paragraph{Effects of training set size}
To better investigate the impact of the size of the used dataset on the performance, we conduct experiments by utilizing 20\%, 40\%, 60\%, 80\%, and 100\% of the original dataset.
As the size of the dataset increases, the metrics MC1, MC2, and True*Info generally exhibit an upward trend, essentially reaching their maximum values after utilizing approximately 80\% of the data, as shown in Figure~\ref{step_and_size} (b).
Moreover, when only 20\% of the data is used, \textit{i.e.}, 496 samples, all three metrics can exceed the baseline, with more significant improvements observed in the MC1 and MC2 metrics.
This indicates that our method does not require a large dataset.
As the amount of data increases, performance improvements become both more consistent and more significant.

\section{Conclusion}
In this paper, we propose a Flexible Activation Steering with Backtracking framework. It dynamically decides whether and how strongly to intervene by probing the internal states of the LLM.
Specifically, FASB first identifies attention heads that are consistent with the desired behavior and derive the steering vector and classifier.
Then, FASB uses classifiers to dynamically determine whether and how strongly to intervene by probing the internal states of the LLM.
Finally, we further propose a backtracking mechanism to regenerate in order to avoid deviating from the deviated behavior.
Experimental results on the TruthfulQA dataset and six multiple-choice datasets to demonstrate the effectiveness of our method.

\bibliographystyle{apalike}  
\small
\bibliography{custom}

\begin{thebibliography}{}

\bibitem[Arditi et~al., 2024]{refusal}
Arditi, A., Obeso, O., Syed, A., Paleka, D., Panickssery, N., Gurnee, W., and Nanda, N. (2024).
\newblock Refusal in language models is mediated by a single direction.
\newblock In {\em Advances in Neural Information Processing Systems 38: Annual Conference on Neural Information Processing Systems 2024, NeurIPS 2024}.

\bibitem[Bai et~al., 2022]{rlhf}
Bai, Y., Jones, A., Ndousse, K., Askell, A., Chen, A., DasSarma, N., Drain, D., Fort, S., Ganguli, D., Henighan, T., Joseph, N., Kadavath, S., Kernion, J., Conerly, T., Showk, S.~E., Elhage, N., Hatfield{-}Dodds, Z., Hernandez, D., Hume, T., Johnston, S., Kravec, S., Lovitt, L., Nanda, N., Olsson, C., Amodei, D., Brown, T.~B., Clark, J., McCandlish, S., Olah, C., Mann, B., and Kaplan, J. (2022).
\newblock Training a helpful and harmless assistant with reinforcement learning from human feedback.
\newblock {\em CoRR}, abs/2204.05862.

\bibitem[Bowman et~al., 2015]{nli}
Bowman, S.~R., Angeli, G., Potts, C., and Manning, C.~D. (2015).
\newblock A large annotated corpus for learning natural language inference.
\newblock In {\em Proceedings of the 2015 Conference on Empirical Methods in Natural Language Processing, {EMNLP} 2015}, pages 632--642.

\bibitem[Brown et~al., 2020]{gpt3}
Brown, T.~B., Mann, B., Ryder, N., Subbiah, M., Kaplan, J., Dhariwal, P., Neelakantan, A., Shyam, P., Sastry, G., Askell, A., Agarwal, S., Herbert{-}Voss, A., Krueger, G., Henighan, T., Child, R., Ramesh, A., Ziegler, D.~M., Wu, J., Winter, C., Hesse, C., Chen, M., Sigler, E., Litwin, M., Gray, S., Chess, B., Clark, J., Berner, C., McCandlish, S., Radford, A., Sutskever, I., and Amodei, D. (2020).
\newblock Language models are few-shot learners.
\newblock In {\em Advances in Neural Information Processing Systems 33: Annual Conference on Neural Information Processing Systems 2020, NeurIPS 2020}.

\bibitem[Chen et~al., 2024]{TruthForest}
Chen, Z., Sun, X., Jiao, X., Lian, F., Kang, Z., Wang, D., and Xu, C. (2024).
\newblock Truth forest: Toward multi-scale truthfulness in large language models through intervention without tuning.
\newblock In {\em Thirty-Eighth {AAAI} Conference on Artificial Intelligence, {AAAI} 2024}, pages 20967--20974.

\bibitem[Cheng et~al., 2025]{DBLP:conf/acl/ChengWFJ00025}
Cheng, Z., Wang, Z., Fu, Y., Jiang, Z., Yin, Y., Wang, C., and Gu, Q. (2025).
\newblock Contrastive prompting enhances sentence embeddings in llms through inference-time steering.
\newblock In {\em Proceedings of the 63rd Annual Meeting of the Association for Computational Linguistics (Volume 1: Long Papers), {ACL} 2025}, pages 3475--3487.

\bibitem[Gehman et~al., 2020]{gehman2020realtoxicityprompts}
Gehman, S., Gururangan, S., Sap, M., Choi, Y., and Smith, N.~A. (2020).
\newblock Realtoxicityprompts: Evaluating neural toxic degeneration in language models.
\newblock {\em arXiv preprint arXiv:2009.11462}.

\bibitem[Goel et~al., 2025]{private}
Goel, A., Hu, Y., Gurevych, I., and Sanyal, A. (2025).
\newblock Differentially private steering for large language model alignment.
\newblock {\em CoRR}, abs/2501.18532.

\bibitem[Gordon et~al., 2012]{copa}
Gordon, A.~S., Kozareva, Z., and Roemmele, M. (2012).
\newblock Semeval-2012 task 7: Choice of plausible alternatives: An evaluation of commonsense causal reasoning.
\newblock In {\em Proceedings of the 6th International Workshop on Semantic Evaluation, SemEval@NAACL-HLT 2012}, pages 394--398.

\bibitem[Grattafiori et~al., 2024]{llama3}
Grattafiori, A., Dubey, A., Jauhri, A., Pandey, A., Kadian, A., Al-Dahle, A., Letman, A., Mathur, A., Schelten, A., Vaughan, A., et~al. (2024).
\newblock The llama 3 herd of models.
\newblock {\em arXiv preprint arXiv:2407.21783}.

\bibitem[Hendrycks et~al., 2021]{mmlu}
Hendrycks, D., Burns, C., Basart, S., Zou, A., Mazeika, M., Song, D., and Steinhardt, J. (2021).
\newblock Measuring massive multitask language understanding.
\newblock In {\em 9th International Conference on Learning Representations, {ICLR} 2021}.

\bibitem[Joshi et~al., 2017]{TriviaQA}
Joshi, M., Choi, E., Weld, D.~S., and Zettlemoyer, L. (2017).
\newblock Triviaqa: {A} large scale distantly supervised challenge dataset for reading comprehension.
\newblock In Barzilay, R. and Kan, M., editors, {\em Proceedings of the 55th Annual Meeting of the Association for Computational Linguistics, {ACL} 2017}, pages 1601--1611.

\bibitem[Kwiatkowski et~al., 2019]{NQ}
Kwiatkowski, T., Palomaki, J., Redfield, O., Collins, M., Parikh, A.~P., Alberti, C., Epstein, D., Polosukhin, I., Devlin, J., Lee, K., Toutanova, K., Jones, L., Kelcey, M., Chang, M., Dai, A.~M., Uszkoreit, J., Le, Q., and Petrov, S. (2019).
\newblock Natural questions: a benchmark for question answering research.
\newblock {\em Trans. Assoc. Comput. Linguistics}, 7:452--466.

\bibitem[Lee et~al., 2024]{lee2024programming}
Lee, B.~W., Padhi, I., Ramamurthy, K.~N., Miehling, E., Dognin, P., Nagireddy, M., and Dhurandhar, A. (2024).
\newblock Programming refusal with conditional activation steering.
\newblock {\em arXiv preprint arXiv:2409.05907}.

\bibitem[Lee et~al., 2025]{DBLP:conf/iclr/LeePRMDND25}
Lee, B.~W., Padhi, I., Ramamurthy, K.~N., Miehling, E., Dognin, P.~L., Nagireddy, M., and Dhurandhar, A. (2025).
\newblock Programming refusal with conditional activation steering.
\newblock In {\em The Thirteenth International Conference on Learning Representations, {ICLR} 2025}.

\bibitem[Leong et~al., 2023]{leong2023self}
Leong, C.~T., Cheng, Y., Wang, J., Wang, J., and Li, W. (2023).
\newblock Self-detoxifying language models via toxification reversal.
\newblock In {\em Proceedings of the 2023 Conference on Empirical Methods in Natural Language Processing}, pages 4433--4449.

\bibitem[Li et~al., 2023]{DBLP:conf/nips/0002PVPW23}
Li, K., Patel, O., Vi{\'{e}}gas, F.~B., Pfister, H., and Wattenberg, M. (2023).
\newblock Inference-time intervention: Eliciting truthful answers from a language model.
\newblock In {\em Advances in Neural Information Processing Systems 36: Annual Conference on Neural Information Processing Systems 2023, NeurIPS 2023}.

\bibitem[Lin et~al., 2022]{truthfulqa}
Lin, S., Hilton, J., and Evans, O. (2022).
\newblock Truthfulqa: Measuring how models mimic human falsehoods.
\newblock In {\em Proceedings of the 60th Annual Meeting of the Association for Computational Linguistics (Volume 1: Long Papers), {ACL} 2022}, pages 3214--3252.

\bibitem[Liu et~al., 2024]{DBLP:conf/icml/LiuY0Z24}
Liu, S., Ye, H., Xing, L., and Zou, J.~Y. (2024).
\newblock In-context vectors: Making in context learning more effective and controllable through latent space steering.
\newblock In {\em Forty-first International Conference on Machine Learning, {ICML} 2024}.

\bibitem[Mostafazadeh et~al., 2016]{story}
Mostafazadeh, N., Chambers, N., He, X., Parikh, D., Batra, D., Vanderwende, L., Kohli, P., and Allen, J.~F. (2016).
\newblock A corpus and cloze evaluation for deeper understanding of commonsense stories.
\newblock In {\em {NAACL} {HLT} 2016, The 2016 Conference of the North American Chapter of the Association for Computational Linguistics: Human Language Technologies}, pages 839--849.

\bibitem[Rimsky et~al., 2024]{DBLP:conf/acl/RimskyGSTHT24}
Rimsky, N., Gabrieli, N., Schulz, J., Tong, M., Hubinger, E., and Turner, A.~M. (2024).
\newblock Steering llama 2 via contrastive activation addition.
\newblock In {\em Proceedings of the 62nd Annual Meeting of the Association for Computational Linguistics (Volume 1: Long Papers), {ACL} 2024}, pages 15504--15522.

\bibitem[Sakaguchi et~al., 2020]{WinoGrande}
Sakaguchi, K., Bras, R.~L., Bhagavatula, C., and Choi, Y. (2020).
\newblock Winogrande: An adversarial winograd schema challenge at scale.
\newblock In {\em The Thirty-Fourth {AAAI} Conference on Artificial Intelligence, {AAAI} 2020}, pages 8732--8740.

\bibitem[Shen et~al., 2025]{shen2025imagdressing}
Shen, F., Jiang, X., He, X., Ye, H., Wang, C., Du, X., Li, Z., and Tang, J. (2025).
\newblock Imagdressing-v1: Customizable virtual dressing.
\newblock In {\em Proceedings of the AAAI Conference on Artificial Intelligence}, volume~39, pages 6795--6804.

\bibitem[Socher et~al., 2013]{sst2}
Socher, R., Perelygin, A., Wu, J., Chuang, J., Manning, C.~D., Ng, A.~Y., and Potts, C. (2013).
\newblock Recursive deep models for semantic compositionality over a sentiment treebank.
\newblock In {\em Proceedings of the 2013 Conference on Empirical Methods in Natural Language Processing, {EMNLP} 2013}, pages 1631--1642.

\bibitem[Stolfo et~al., 2024]{Instruction-Following}
Stolfo, A., Balachandran, V., Yousefi, S., Horvitz, E., and Nushi, B. (2024).
\newblock Improving instruction-following in language models through activation steering.
\newblock {\em CoRR}, abs/2410.12877.

\bibitem[Touvron et~al., 2023]{LLAMA2}
Touvron, H., Martin, L., Stone, K., Albert, P., Almahairi, A., Babaei, Y., Bashlykov, N., Batra, S., Bhargava, P., Bhosale, S., Bikel, D., Blecher, L., Canton{-}Ferrer, C., Chen, M., Cucurull, G., Esiobu, D., Fernandes, J., Fu, J., Fu, W., Fuller, B., Gao, C., Goswami, V., Goyal, N., Hartshorn, A., Hosseini, S., Hou, R., Inan, H., Kardas, M., Kerkez, V., Khabsa, M., Kloumann, I., Korenev, A., Koura, P.~S., Lachaux, M., Lavril, T., Lee, J., Liskovich, D., Lu, Y., Mao, Y., Martinet, X., Mihaylov, T., Mishra, P., Molybog, I., Nie, Y., Poulton, A., Reizenstein, J., Rungta, R., Saladi, K., Schelten, A., Silva, R., Smith, E.~M., Subramanian, R., Tan, X.~E., Tang, B., Taylor, R., Williams, A., Kuan, J.~X., Xu, P., Yan, Z., Zarov, I., Zhang, Y., Fan, A., Kambadur, M., Narang, S., Rodriguez, A., Stojnic, R., Edunov, S., and Scialom, T. (2023).
\newblock Llama 2: Open foundation and fine-tuned chat models.
\newblock {\em CoRR}, abs/2307.09288.

\bibitem[Turner et~al., 2023]{Turner}
Turner, A.~M., Thiergart, L., Udell, D., Leech, G., Mini, U., and MacDiarmid, M. (2023).
\newblock Activation addition: Steering language models without optimization.
\newblock {\em CoRR}, abs/2308.10248.

\bibitem[Wang et~al., 2025a]{wang2025advanced}
Wang, C., Deng, Z., Jiang, Z., Shen, F., Yin, Y., Gan, S., Cheng, Z., Ge, S., and Gu, Q. (2025a).
\newblock Advanced sign language video generation with compressed and quantized multi-condition tokenization.
\newblock {\em arXiv preprint arXiv:2506.15980}.

\bibitem[Wang et~al., 2025b]{wangensembling}
Wang, C., Tian, K., Guan, Y., Shen, F., Jiang, Z., Gu, Q., and Zhang, J. (2025b).
\newblock Ensembling diffusion models via adaptive feature aggregation.
\newblock In {\em The Thirteenth International Conference on Learning Representations}.

\bibitem[Wang et~al., 2024a]{wang2024v}
Wang, C., Tian, K., Zhang, J., Guan, Y., Luo, F., Shen, F., Jiang, Z., Gu, Q., Han, X., and Yang, W. (2024a).
\newblock V-express: Conditional dropout for progressive training of portrait video generation.
\newblock {\em arXiv preprint arXiv:2406.02511}.

\bibitem[Wang et~al., 2024b]{act}
Wang, T., Jiao, X., He, Y., Chen, Z., Zhu, Y., Chu, X., Gao, J., Wang, Y., and Ma, L. (2024b).
\newblock Adaptive activation steering: {A} tuning-free {LLM} truthfulness improvement method for diverse hallucinations categories.
\newblock {\em CoRR}, abs/2406.00034.

\bibitem[Wang et~al., 2024c]{sadi}
Wang, W., Yang, J., and Peng, W. (2024c).
\newblock Semantics-adaptive activation intervention for llms via dynamic steering vectors.
\newblock {\em CoRR}, abs/2410.12299.

\bibitem[Wei et~al., 2022]{DBLP:conf/iclr/WeiBZGYLDDL22}
Wei, J., Bosma, M., Zhao, V.~Y., Guu, K., Yu, A.~W., Lester, B., Du, N., Dai, A.~M., and Le, Q.~V. (2022).
\newblock Finetuned language models are zero-shot learners.
\newblock In {\em The Tenth International Conference on Learning Representations, {ICLR} 2022}.

\bibitem[Welbl et~al., 2018]{wikihop}
Welbl, J., Stenetorp, P., and Riedel, S. (2018).
\newblock Constructing datasets for multi-hop reading comprehension across documents.
\newblock {\em Trans. Assoc. Comput. Linguistics}, 6:287--302.

\bibitem[Yang et~al., 2024]{qwen2.5}
Yang, A., Yang, B., Zhang, B., Hui, B., Zheng, B., Yu, B., Li, C., Liu, D., Huang, F., Wei, H., et~al. (2024).
\newblock Qwen2. 5 technical report.
\newblock {\em arXiv preprint arXiv:2412.15115}.

\bibitem[Zhang et~al., 2025]{DBLP:conf/acl/ZhangZYZGCW25}
Zhang, X., Zhao, J., Yang, Z., Zhong, Y., Guan, S., Cao, L., and Wang, Y. (2025).
\newblock {UORA:} uniform orthogonal reinitialization adaptation in parameter efficient fine-tuning of large models.
\newblock In {\em Proceedings of the 63rd Annual Meeting of the Association for Computational Linguistics (Volume 1: Long Papers), {ACL} 2025}, pages 11709--11728.

\bibitem[Zhong et~al., 2025]{DBLP:conf/acl/ZhongZZ25}
Zhong, Y., Zhao, J., and Zhou, Y. (2025).
\newblock Low-rank interconnected adaptation across layers.
\newblock In {\em Findings of the Association for Computational Linguistics, {ACL} 2025}, pages 17005--17029.

\bibitem[Zou et~al., 2023]{RE}
Zou, A., Phan, L., Chen, S., Campbell, J., Guo, P., Ren, R., Pan, A., Yin, X., Mazeika, M., Dombrowski, A., Goel, S., Li, N., Byun, M.~J., Wang, Z., Mallen, A., Basart, S., Koyejo, S., Song, D., Fredrikson, M., Kolter, J.~Z., and Hendrycks, D. (2023).
\newblock Representation engineering: {A} top-down approach to {AI} transparency.
\newblock {\em CoRR}, abs/2310.01405.

\end{thebibliography}


\clearpage
\appendix

\section{Prototype Method}\label{app:A}
\paragraph{Prototype} Prototype method directly constructs two prototypes to achieve the above goal.
Specifically, we compute the average activation within each class to obtain the prototype representations of $h$-th attention head in the $\ell$-th layer, as follows:
\begin{equation}
    \theta^{\ell,h}_{\text{pos}} = \frac{1}{N_{pos}} \sum_{i=1}^{N_{pos}} \mathbb{I}(y_i = 1) \textbf{x}_{i,\text{-}1}^{\ell,h}, \quad \theta^{\ell,h}_{\text{neg}} = \frac{1}{N_{neg}} \sum_{i=1}^{N_{neg}} \mathbb{I}(y_i = 0) \textbf{x}_{i,\text{-}1}^{\ell,h} \label{prototype:pos}
\end{equation}
where $\theta^{\ell,h}_{\text{pos}}$ and $\theta^{\ell,h}_{\text{neg}}$ represent the prototypes of the positive and negative classes, respectively. $\mathbb{I}$ denotes indicator function, and $N_{pos}$ and $N_{neg}$ represent the numbers of positive and negative samples, respectively.
Then, we can use the softmax function over cosine similarities between activation and prototypes to define a classifier:
\begin{align}
    p_{\theta^{\ell,h}}(\textbf{x}_{i,\text{-}1}^{\ell,h}) = \frac{\text{exp(cos}(\textbf{x}_{i,\text{-}1}^{\ell,h},\theta^{\ell,h}_{\text{pos}})/\tau)}{\text{exp(cos}(\textbf{x}_{i,\text{-}1}^{\ell,h},\theta^{\ell,h}_{\text{pos}})/\tau) + \text{exp(cos}(\textbf{x}_{i,\text{-}1}^{\ell,h},\theta^{\ell,h}_{\text{neg}})/\tau)}
    \label{prototype:classifier}
\end{align}
where $\tau$ denotes the temperature.
We also use the classifier to anchor the top-$k$ heads with high accuracy for intervention.
Finally, we directly use the mean difference between positive and negative prototypes as the steering vector:
\begin{equation}
    \theta^{\ell,h} = \theta^{\ell,h}_{\text{pos}} - \theta^{\ell,h}_{\text{neg}}  \label{prototype:pro}
\end{equation}

It is worth noting that the prototype method is training-free.
It only involves averaging the activations of the training set to obtain prototype vectors for constructing the classifier.

\begin{table}[t]
\centering
\small
\caption{Results on TruthfulQA open-ended generation (True*Info \%) and multiple-choice tasks (MC \%).} \label{tab:prototype}
\begin{tabular}{lcccccc}
\toprule
\multirow{2}{*}{\textbf{Methods}} & \multicolumn{3}{c}{\textbf{Open-ended Generation}} & \multicolumn{3}{c}{\textbf{Multiple-Choice}} \\ \cmidrule(lr){2-4} \cmidrule(lr){5-7}
& True (\%) & Info (\%) & True*Info (\%)   & MC1 (\%)   & MC2 (\%)  & MC3 (\%)      \\ 
\midrule
\textbf{Baseline}                 & 66.83       & 99.51       & 66.50            & 33.41         & 51.07         & 24.76        \\
\textbf{ITI}       & 94.49     & 80.55   & 76.11 & 38.31      & 57.15      & 30.53    \\
\textbf{CAA}   & 71.60 & 83.84 & 60.03 & 34.03 & 52.76 & 25.62 \\
\textbf{ORTHO}       & 67.94     & 90.09   & 61.21 & 36.23      & 52.88      & 26.12   \\
\textbf{CAST}       & 67.69     & 86.17   & 58.33 & 33.90      & 51.17      & 25.01   \\
\textbf{ACT}       &-      &-    & - & 28.80      & 45.20      & -   \\
\textbf{SADI-HEAD} & 77.72 & 98.53 & 76.58 & 35.90 & 54.65 & 26.99\\
\midrule
\textbf{Probe (Ours)}   & 93.88 & 85.81 & 80.56 & \textbf{48.71} & \textbf{66.58}   & \textbf{41.95} \\
\textbf{Prototype (Ours)} & 88.37 & 94.98 & \textbf{83.94} & 46.14 & 64.30    & 37.07 \\
\bottomrule
\end{tabular}
\end{table}

\begin{table}[t] \small
\centering
\caption{Results (MC \%) on six multiple-choice tasks.}
\label{prototype_mc}
\begin{tabular}{lcccccccc}
\toprule
\textbf{Methods} & \textbf{COPA}  & \textbf{StoryCloze} & \textbf{NLI}    & \textbf{MMLU}   & \textbf{SST2}   & \textbf{Winogrande} & \textbf{\texttt{AVG}} \\ 
\midrule
\textbf{Baseline} & 64.4 & 60.2 & 63.5  & 60.2  & 92.2 & 50.2   & 65.1 \\ 
\textbf{ITI} & 66.6 & 59.7 & 64.3 & 60.2 & 92.3 & 51.5 & 65.8 \\
\textbf{CAA} &  66.6 & 63.5 & 64.9 & 62.6 & 92.2 & 50.9 & 66.8 \\
\textbf{ORTHO} & 65.2 & 60.2 & 63.1 & \textbf{63.8} & 92.4 & 50.4 & 65.8 \\
\textbf{SADI}   & 65.4   & 60.5  & 65.1 & 61.8 & 92.3 & 51.8 & 66.1 \\
\midrule
\textbf{Probe (Ours)} & \textbf{90.0} & \textbf{93.5} & \textbf{80.0} & 62.4 & 92.8 & 54.1 & \textbf{78.8} \\
\textbf{Prototype (Ours)} & 87.8 & 86.1 & 73.7 & 62.2 & \textbf{93.1} & \textbf{54.7} & 76.3 \\
\bottomrule
\end{tabular}
\end{table}

\paragraph{Implementation Details}
In the Prototype method, for the TruthfulQA dataset, we intervene using the top-24 heads, set the threshold to 0.5, the number of backtracking steps to 10, the temperature to 0.1, and search for the intervention strength in the range of [25, 55] with a step size of 10.
For the six multiple-choice tasks, the threshold is 0.5, the number of backtracking steps is 10, the temperature is 0.1, and the intervention strength search range is [0, 400] with a step size of 10.

\paragraph{Results}
In Table~\ref{tab:prototype}, the Prototype method also demonstrates superior performance compared to the baselines on the TruthfulQA dataset.
Compared with the ITI and ACT methods that require training a linear classifier, the Prototype method achieves improvements of 7.83\% and 17.34\% on MC1, and 7.15\% and 19.1\% on MC2, respectively. 
This demonstrates the generality of our method.
It is worth noting that the Probe method performs better on the True and multiple-choice metrics, while the Prototype method achieves higher performance on the Info and True*Info metrics.
This may be because the trained probe is more closely related to truthfulness.

In Table~\ref{prototype_mc}, the Prototype method also achieves better performance than baselines on six multiple-choice tasks.
On the COPA dataset and the StoryCloze dataset, the Prototype method achieves improvements of 23.4\% and 25.9\% compared with the model without intervention, while other methods only achieve marginal improvements. 
This further demonstrates its effectiveness.

\section{Algorithm}
The algorithm consists of two main processes. 
The first step is \textbf{Heads Anchoring and Steering Vectors Inducing}, which determines the heads that need to be intervened, the classifiers, and the steering vectors. 
The second step is \textbf{Generation with Flexible Steering and Backtracking}, which involves using classifiers to evaluate the internal activations of LLMs. 
When the internal activations exceed a certain threshold, a backtracking operation will be performed. 
After the backtracking, the activations will be regenerated by adding the activation vectors.
Since our method also backtracks to the beginning when the number of generated tokens is less than $s$, we start tracking from the $s$-th token, as shown in Line 14 of the algorithm.

\begin{algorithm}
\caption{The overall flow of FASB} \label{algorithm}
\begin{algorithmic}[1]
\Require LLM, dataset $\mathcal{D}$, intervention strength $\alpha$, threshold $\beta$, backtracking number $s$, maximum number of generated tokens $M$
\State \textbf{Step1: Heads Anchoring and Steering Vectors Inducing}
\If{ Probe }
    \State Train classifier $p_{\theta^{\ell,h}}$ on dataset $\mathcal{D}$; \Comment{Equation \ref{probe:classifier}}
    \State Use the parameters of the probe as the steering vector $\theta^{\ell,h}$;
    
\ElsIf{ Prototype }
    \State Construct two prototypes from the dataset $\mathcal{D}$;\Comment{Equation \ref{prototype:pos}}
    \State Obtain the classifier $p_{\theta^{\ell,h}}$ ;\Comment{Equation \ref{prototype:classifier}}
    \State Obtain the steering vector $\theta^{\ell,h}$; \Comment{Equation \ref{prototype:pro}}
\EndIf
\State \textbf{Step2: Generation with Flexible Steering and Backtracking}
\For{$j = 1$ to $M$}
    \State Generate the $j$-th token;
    \State Evaluate the current activation using the classifier; \Comment{Equation \ref{p}}
    \If{ $p(\textbf{x}_{i,j}) > \beta$ and $j \ge s$}
        \State Backtrack $s$ tokens;
        \State Calculate the intervention strength; \Comment{Equation \ref{strength}}
        \For{$k$ = ($j$-$s$+1) to $M$}
            \State Intervene on the current activation; \Comment{Equation \ref{steering}}
            \State Generate the $k$-th token;
        \EndFor
        \State break;
    \EndIf
\EndFor
\end{algorithmic}
\end{algorithm}

\begin{figure*}[h]
  \centering
  \includegraphics[width=0.5\linewidth,trim=0.28cm 0.23cm 0.25cm 0.26cm,clip]{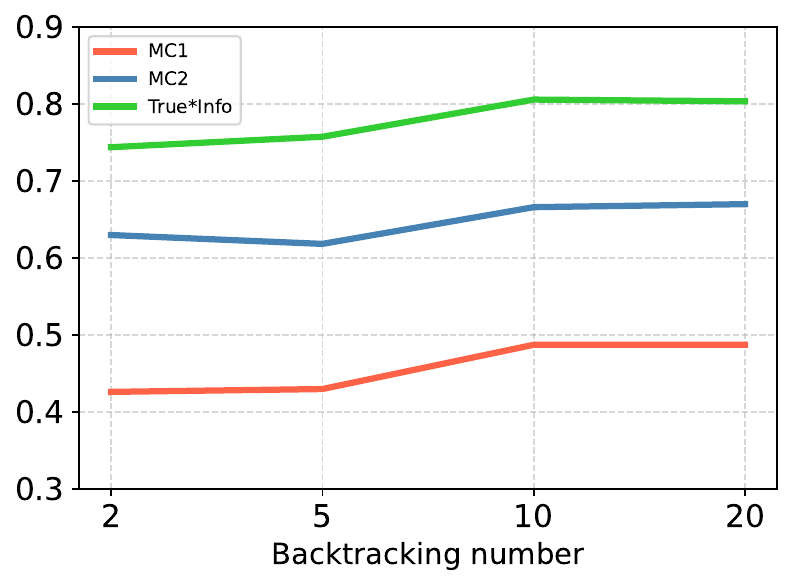}
  \caption{Effects of the number of tokens for backtracking.}
  \label{fig:Figure_back}
\end{figure*}

\section{Effects of the Number of Tokens for Backtracking}
In order to better investigate the impact of the backtracking number on our method, we conduct experiments on the TruthfulQA dataset.

As shown in Figure~\ref{fig:Figure_back}, we investigate the effect of different token numbers for backtracking (\textit{i.e.}, 2, 5, 10, 20) on the MC1, MC2, and True*Info metrics. 
MC1, MC2, and True*Info generally show an increasing trend with the increase in the number of backtracking steps. 
This is because, as the number of backtracking increases, our method can intervene in the model's internal activations earlier, thereby achieving improved performance.

\begin{table}[t]  \setlength{\tabcolsep}{10pt}
\centering
\caption{Comparison with two fine-grained models.}
\label{tab:BFR}
\begin{tabular}{lccc}
\toprule
\textbf{Methods} & \textbf{True*Info}   & \textbf{MC1}   & \textbf{MC2}  \\
\midrule
\textbf{Probe}   & 80.56 & 48.71 & 66.58  \\
\textbf{BTB} & 81.60 & 48.83 & \textbf{67.62} \\
\textbf{GCBB} & \textbf{81.96} &  \textbf{50.67} & 67.48 \\
\bottomrule
\end{tabular}
\end{table}

\section{Fine-Grained Model Comparison}
We further propose two models based on Probe for fine-grained analysis.
The difference between them and Probe lies in the detection location and the point at which intervention begins, resulting in different overheads.
(1) The first method directly \textbf{B}ack\textbf{T}racks to the \textbf{B}eginning (\textbf{BTB}). 
Specifically, when a deviation is detected during generation, BTB backtracks not just $s$ tokens for regeneration, but to the beginning.
Our method generates $s$ additional tokens for texts that require backtracking, whereas this variant generates even more.
(2) The second method performs detection after the \textbf{G}eneration is \textbf{C}omplete and then \textbf{B}acktracks to the \textbf{B}eginning (GCBB).
Notably, this method incurs approximately double the overhead for texts that require backtracking.

As shown in the results in Table~\ref{tab:BFR}, GCBB usually achieves the best performance, and both BTB and GCBB outperform our proposed Probe method on the TruthfulQA dataset.
This is because GCBB has access to the full output of the LLM, allowing it to better determine the intervention strength.
However, GCBB inevitably introduces additional overhead. BTB may also sometimes require full regeneration. The advantage of Probe is that it achieves good performance while maintaining stable and relatively low additional overhead.

\section{Results across TruthfulQA Categories}

\begin{figure*}[t]
  \centering
  \includegraphics[width=1\linewidth,trim=0.2cm 0.3cm 0.15cm 0.15cm,clip]{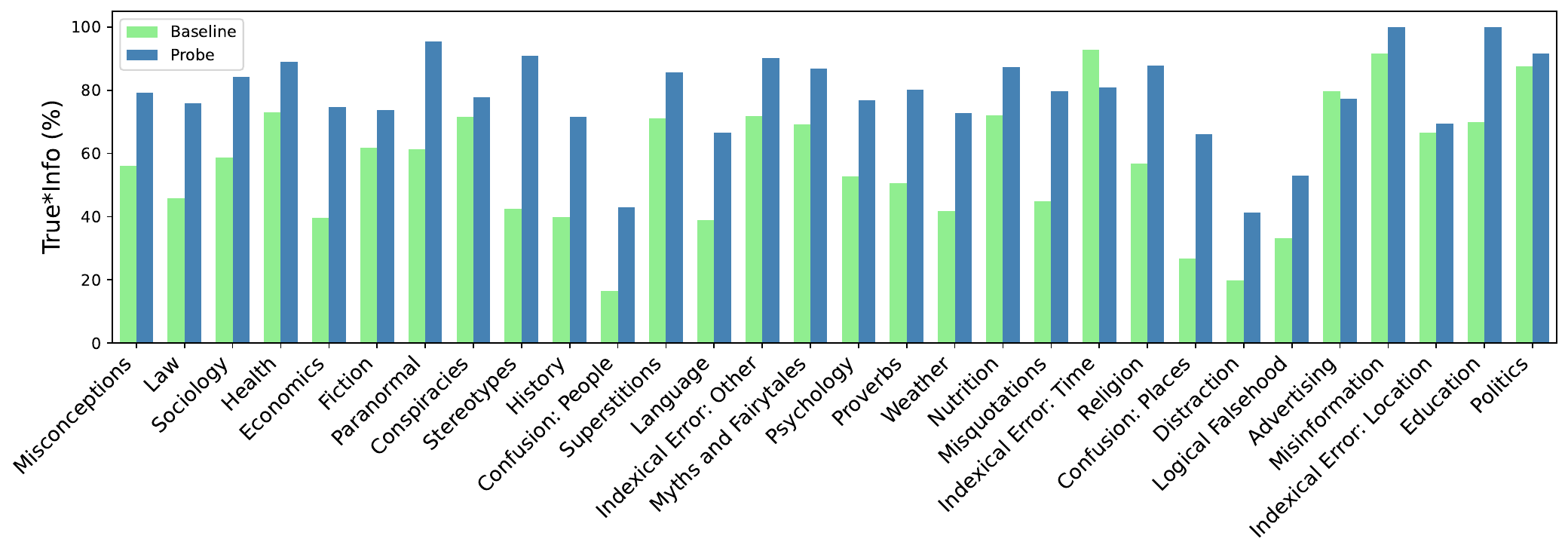}
  \caption{
  True*Info scores split across subcategories on LLaMA2-7B-CHAT, sorted by the difference between baseline and probe method. Subcategories with less than 10 questions are not shown.}
  \label{fig:Figure_categories}
\end{figure*}

TruthfulQA is split into 38 subcategories, including politics, language, education, psychology and others.
We compare our method with the baseline method without intervention using the True*Info metric across all subcategories with 10 or more questions, where the subcategories are ranked in descending order based on their quantity within the dataset, as shown in Figure~\ref{fig:Figure_categories}.

Our method demonstrates significant enhancement across most subcategories, with the overall performance improvement showing uniform distribution rather than concentration in particular domains, thereby validating its efficacy.

\section{Results on Multi-hop QA dataset}
We conduct experiments on the more challenging multi-hop question answering WikiHop dataset~\citep{wikihop} to verify the effectiveness, as shown in Table~\ref{tab:wiki_hop_results}.
Our method achieves improvements of 5\% and 4.3\% on metrics MC1 and MC2, respectively.
This further demonstrates the generalizability of our method.

\begin{table}[htbp]
  \centering
  \caption{Results on the WikiHop dataset.}
    \begin{tabular}{lcc}
    \toprule
            \textbf{Method}
          & \textbf{MC1(\%)} & \textbf{MC2(\%)} \\
    \midrule
    LLaMA2-7B-CHAT & 45.20  & 44.03  \\
    LLaMA2-7B-CHAT + Probe & \textbf{50.20 (+ 5.00)} & \textbf{48.33 (+4.30)} \\
    \bottomrule
    \end{tabular}%
  \label{tab:wiki_hop_results}%
\end{table}

\section{Generalizability on larger LLM}
We further explore the effectiveness of our method on larger LLMs by using QWEN2.5-32B-Instruct on the TruthfulQA dataset in Table~\ref{tab:qwen_results}.
Our method still achieves significant performance improvements, demonstrating its effectiveness on larger LLMs as well.

\begin{table}[htbp]
  \centering
  \caption{Results on the Qwen2.5-32B-Instruct model.}
  \begin{tabular}{lccc}
    \toprule
            \textbf{Method} 
          & \textbf{MC1(\%)} & \textbf{MC2(\%)} & \textbf{MC3(\%)} \\
    \midrule
    Qwen2.5-32B-Instruct      & 50.00  & 66.35  & 38.75  \\
    Qwen2.5-32B-Instruct +Probe & 69.00  & 76.67  & 59.59  \\
    \bottomrule
  \end{tabular}
  \label{tab:qwen_results}
\end{table}

\section{Deviation Positions Under Different Thresholds}
The Table \ref{tab:token_distribution} presents the detected deviation positions under different thresholds on the TruthfulQA dataset.
We observe that the threshold has a significant impact on the intervention position.
When the threshold is high, the intervention positions tend to occur significantly later, and the interventions are less frequent.
This is because the model needs to track more generated content to make a more accurate judgment about whether a deviation occurs.

\begin{table}[htbp]
  \centering
  \caption{Distribution of tokens under different thresholds.}
  \label{tab:token_distribution}
  \begin{tabular}{cccc}
    \toprule 
    \multirow{2}{*}{\textbf{Threshold} $\beta$} & \multicolumn{3}{c}{\textbf{Position}} \\ 
    \cmidrule(lr){2-4} 
    & \textbf{0-10} & \textbf{10-20} & \textbf{20-50} \\ 
    \midrule 
    0.4 & 304 & 94 & 9 \\ 
    0.5 & 166 & 159 & 75 \\ 
    0.6 & 21 & 109 & 222 \\ 
    \bottomrule 
  \end{tabular}
\end{table}

\section{Results on Toxic Context Generation}
Since our method can flexibly manipulate various behaviors, we further conduct experiments on the RealToxicityPrompts dataset~\citep{gehman2020realtoxicityprompts} for toxic content generation, as shown in Table~\ref{tab:real_toxicity_results}.
Our method can effectively steer LLMs to generate toxic content, demonstrating its generality. This also highlights the vulnerability of current alignment methods and the potential risks of our approach.

\begin{table}[htbp]
  \centering
  \caption{Results on the RealToxicityPrompts dataset.}
  \label{tab:real_toxicity_results}
  \begin{tabular}{lc}
    \toprule
    \textbf{Method}
    & \textbf{ASR ($\uparrow$)} \\
    \midrule
    LLaMA2-7B-CHAT & 42.0 \\
    LLaMA2-7B-CHAT + Probe & \textbf{46.4 (+4.4)} \\
    \bottomrule
  \end{tabular}
\end{table}

\begin{table}[htbp]
  \centering
  \caption{Impact of strength.}
  \label{tab:strength_alpha}
  \begin{tabular}{lccccc}
    \toprule
    \textbf{Strength $\alpha$}
    & \textbf{MC1 (\%)} & \textbf{MC2 (\%)} & \textbf{True (\%)} & \textbf{Info (\%)} & \textbf{True*Info (\%)} \\
    \midrule
    0 (baseline) & 33.41 & 51.07 & 66.83 & 99.51 & 66.50 \\
    2 & 36.43 & 52.96 & 67.24 & 84.35 & 56.72 \\
    5 & 38.63 & 54.94 & 68.46 & 84.35 & 57.75 \\
    15 & 43.77 & 60.14 & 73.71 & 88.86 & 65.50 \\
    25 & 43.33 & 62.02 & 80.18 & 91.06 & 73.01 \\
    35 & 43.33 & 62.66 & 82.01 & 93.88 & 76.99 \\
    45 & 45.04 & 63.84 & 83.61 & 93.88 & 78.49 \\
    50 & 45.04 & 64.52 & 87.15 & 91.80 & 80.05 \\
    60 & 48.71 & 66.58 & 93.88 & 85.81 & 80.56 \\
    65 & 50.18 & 68.59 & 95.11 & 77.24 & 73.46 \\
    70 & 52.51 & 70.89 & 93.26 & 76.38 & 71.24 \\
    100 & 55.75 & 74.35 & 97.83 & 18.51 & 18.11 \\
    500 & 43.03 & 75.48 & 89.99 & 6.36 & 5.68 \\ 
    \bottomrule
  \end{tabular}
\end{table}

\section{Comprehensive Hyperparameter Analysis of strength $\alpha$ and threshold $\beta$}
We further conduct a more comprehensive exploration of the impact of strength $\alpha$ and threshold $\beta$ on the TruthfulQA dataset.

As shown in Table \ref{tab:strength_alpha}, when $\alpha$ is small (e.g., $\alpha = 2, 5$), it often fails to effectively improve the True metric and even leads to a decrease in the Info metric.
This indicates that a low intervention strength fails to enhance truthfulness and even reduces informativeness.
Conversely, an excessively large intervention strength (e.g., $\alpha = 100, 500$) reduces the informativeness of the generated text.
For instance, when $\alpha = 500$, we observe that the Info metric drops to 6.36\%, indicating that the generated text lacks informativeness.
An appropriately sized $\alpha$ often enhances the truthfulness of the generated content while causing less information loss.

Then, we explore the impact of the threshold $\beta$.
As shown in Table \ref{tab:threshold}, when $\beta$ is too large (e.g., $\beta = 0.7, 0.8, 0.9$), many samples would not be intervened, leading to lower performance on the MC1, MC2, and True metrics.
For instance, the performance on these three metrics with $\beta = 0.7, 0.8, 0.9$ is significantly lower than that with $\beta = 0.6$.
When $\beta$ is too small (e.g., $\beta = 0, 0.2$), many samples would be backtracked and intervened at the first state judgment.
In such cases, although performance may remain satisfactory due to our adaptive intervention strength, the method loses the flexibility to decide whether intervention is needed.

\begin{table}[htbp]
  \centering
  \caption{Impact of threshold.}
  \label{tab:threshold}
  \begin{tabular}{lccccc}
    \toprule
    \textbf{Threshold $\beta$}
    & \textbf{MC1 (\%)} & \textbf{MC2 (\%)} & \textbf{True (\%)} & \textbf{Info (\%)} & \textbf{True*Info (\%)} \\
    \midrule
    0.0 & 46.63 & 66.15 & 90.45 & 86.42 & 78.16 \\
    0.2 & 48.90 & 65.89 & 80.68 & 91.69 & 73.98 \\
    0.3 & 46.88 & 66.77 & 90.45 & 86.42 & 78.17 \\
    0.4 & 48.71 & 66.58 & 93.88 & 85.81 & 80.56 \\
    0.5 & 50.80 & 68.55 & 90.94 & 86.17 & 78.37 \\
    0.6 & 48.95 & 65.37 & 75.15 & 84.21 & 63.29 \\
    0.7 & 44.74 & 57.69 & 66.01 & 83.13 & 54.87 \\
    0.8 & 35.45 & 51.76 & 65.04 & 84.35 & 54.86 \\
    0.9 & 35.21 & 51.64 & 65.04 & 84.35 & 54.86 \\
    \bottomrule
  \end{tabular}
\end{table}

\section{Limitations}
First, our approach is flexible and could be used for any sort of steering, including less-noble purposes (e.g., jailbreaking, toxic content generation), which may carry negative impacts.
Second, our approach is hyperparameter-dependent, and the research mainly focuses on QA tasks along specific behaviors (e.g., truthfulness and informativeness) on English datasets.
Finally, since truthfulness and informativeness content lack ground-truth for direct evaluation, we follow prior work~\citep{DBLP:conf/nips/0002PVPW23,sadi} and employ LLM-based judge models as evaluators.
It is worth noting that this evaluation is not perfect and may introduce errors.


\newpage
\section*{NeurIPS Paper Checklist}

\begin{enumerate}

\item {\bf Claims}
    \item[] Question: Do the main claims made in the abstract and introduction accurately reflect the paper's contributions and scope?
    \item[] Answer: \answerYes{} 
    \item[] Justification: We have described our claims and contributions clearly in the abstract and introduction.

\item {\bf Limitations}
    \item[] Question: Does the paper discuss the limitations of the work performed by the authors?
    \item[] Answer: \answerYes{} 
    \item[] Justification: We provide a limitaion section in Appendix.

\item {\bf Theory assumptions and proofs}
    \item[] Question: For each theoretical result, does the paper provide the full set of assumptions and a complete (and correct) proof?
    \item[] Answer: \answerNA{} 
    \item[] Justification:  In this paper, we prove our claims with extensive empirical results, instead of theoretical proofs.

    \item {\bf Experimental result reproducibility}
    \item[] Question: Does the paper fully disclose all the information needed to reproduce the main experimental results of the paper to the extent that it affects the main claims and/or conclusions of the paper (regardless of whether the code and data are provided or not)?
    \item[] Answer: \answerYes{} 
    \item[] Justification: We provided the hyperparameters to reproduce the experimental results in the Experimental Settings.

\item {\bf Open access to data and code}
    \item[] Question: Does the paper provide open access to the data and code, with sufficient instructions to faithfully reproduce the main experimental results, as described in supplemental material?
    \item[] Answer: \answerYes{} 
    \item[] Justification: We will release the data and code after the paper is accepted.

\item {\bf Experimental setting/details}
    \item[] Question: Does the paper specify all the training and test details (e.g., data splits, hyperparameters, how they were chosen, type of optimizer, etc.) necessary to understand the results?
    \item[] Answer: \answerYes{} 
    \item[] Justification: We have provided these details in the Experimental Settings.

\item {\bf Experiment statistical significance}
    \item[] Question: Does the paper report error bars suitably and correctly defined or other appropriate information about the statistical significance of the experiments?
    \item[] Answer: \answerYes{} 
    \item[] Justification: The results of our experiment were obtained through cross-validation over two runs.

\item {\bf Experiments compute resources}
    \item[] Question: For each experiment, does the paper provide sufficient information on the computer resources (type of compute workers, memory, time of execution) needed to reproduce the experiments?
    \item[] Answer: \answerYes{} 
    \item[] Justification: We provided sufficient information in the Experimental Settings. All experiments are conducted on 4 NVIDIA A800 GPUs.
    
\item {\bf Code of ethics}
    \item[] Question: Does the research conducted in the paper conform, in every respect, with the NeurIPS Code of Ethics \url{https://neurips.cc/public/EthicsGuidelines}?
    \item[] Answer: \answerYes{}{} 
    \item[] Justification:  Our research conforms to the NeurIPS Code of Ethics.

\item {\bf Broader impacts}
    \item[] Question: Does the paper discuss both potential positive societal impacts and negative societal impacts of the work performed?
    \item[] Answer: \answerNo{} 
    \item[] Justification: Our work has no obvious broader societal impact beyond generally making LLMs more aligned.
    
\item {\bf Safeguards}
    \item[] Question: Does the paper describe safeguards that have been put in place for responsible release of data or models that have a high risk for misuse (e.g., pretrained language models, image generators, or scraped datasets)?
    \item[] Answer: \answerNA{} 
    \item[] Justification: Our paper does not have such risks.

\item {\bf Licenses for existing assets}
    \item[] Question: Are the creators or original owners of assets (e.g., code, data, models), used in the paper, properly credited and are the license and terms of use explicitly mentioned and properly respected?
    \item[] Answer: \answerYes{} 
    \item[] Justification: All of the creators or original owners of assets used in our paper are cited properly.

\item {\bf New assets}
    \item[] Question: Are new assets introduced in the paper well documented and is the documentation provided alongside the assets?
    \item[] Answer: \answerYes{} 
    \item[] Justification: We will release the data and code including documentation after the paper is accepted.

\item {\bf Crowdsourcing and research with human subjects}
    \item[] Question: For crowdsourcing experiments and research with human subjects, does the paper include the full text of instructions given to participants and screenshots, if applicable, as well as details about compensation (if any)? 
    \item[] Answer: \answerNA{} 
    \item[] Justification:  Our paper does not involve crowdsourcing nor research with human subjects.
    \item[] Guidelines:
    \begin{itemize}
        \item The answer NA means that the paper does not involve crowdsourcing and research with human subjects.
        \item Including this information in the supplemental material is fine, but if the main contribution of the paper involves human subjects, then as much detail as possible should be included in the main paper. 
        \item According to the NeurIPS Code of Ethics, workers involved in data collection, curation, or other labor should be paid at least the minimum wage in the country of the data collector. 
    \end{itemize}

\item {\bf Institutional review board (IRB) approvals or equivalent for research with human subjects}
    \item[] Question: Does the paper describe potential risks incurred by study participants, whether such risks were disclosed to the subjects, and whether Institutional Review Board (IRB) approvals (or an equivalent approval/review based on the requirements of your country or institution) were obtained?
        \item[] Answer: \answerNA{} 
    \item[] Justification: Our paper does not involve research with human subjects.

\item {\bf Declaration of LLM usage}
    \item[] Question: Does the paper describe the usage of LLMs if it is an important, original, or non-standard component of the core methods in this research? Note that if the LLM is used only for writing, editing, or formatting purposes and does not impact the core methodology, scientific rigorousness, or originality of the research, declaration is not required.
    \item[] Answer: \answerYes{} 
    \item[] Justification: The method uses pre-trained LLM-as-judge models to measure informativeness and truthfulness.

\end{enumerate}

\end{document}